\documentclass[pdflatex,sn-mathphys]{sn-jnl}
\usepackage{diagbox} 
\usepackage{color}

\jyear{2021}%

\theoremstyle{thmstyleone}%
\theoremstyle{thmstyletwo}%

\theoremstyle{thmstylethree}%

\begin{document}

\title[FindVehicle and VehicleFinder]{FindVehicle and VehicleFinder: A NER dataset for natural language-based vehicle retrieval and a keyword-based cross-modal vehicle retrieval system}


\author[1,2,3,4]{\fnm{Runwei} \sur{Guan}}\email{Runwei.Guan@liverpool.ac.uk}
\equalcont{These authors contributed equally to this work.}

\author[2]{\fnm{Ka Lok} \sur{Man}}\email{Ka.Man@xjtlu.edu.cn}
\equalcont{These authors contributed equally to this work.}

\author[2]{\fnm{Feifan} \sur{Chen}}\email{sgfchen5@liverpool.ac.uk}
\equalcont{These authors contributed equally to this work.}

\author[1,2,3,4]{\fnm{Shanliang} \sur{Yao}}\email{shanliang.yao@liverpool.ac.uk}

\author[6]{\fnm{Rongsheng} \sur{Hu}}\email{1033170432@stu.jiangnan.edu.cn}

\author[2]{\fnm{Xiaohui} \sur{Zhu}}\email{Xiaohui.Zhu@xjtlu.edu.cn}

\author[1]{\fnm{Jeremy} \sur{Smith}}\email{J.S.Smith@liverpool.ac.uk}

\author[2]{\fnm{Eng Gee} \sur{Lim}}\email{enggee.lim@xjtlu.edu.cn}

\author*[3,4,5]{\fnm{Yutao} \sur{Yue}}\email{yueyutao@idpt.org}

\affil[1]{\orgdiv{Department of Electrical Engineering and Electronics}, \orgname{University of Liverpool}, \orgaddress{
\city{Liverpool}, \postcode{L69 3BX}, 
\country{United Kingdom}}}

\affil[2]{\orgdiv{School of Advanced Technology}, \orgname{Xi'an Jiaotong-Liverpool University}, 
\orgaddress{
\city{Suzhou}, \postcode{215123}, 
\country{China}}}

\affil[3]{\orgdiv{XJTLU-JITRI Academy of Technology}, \orgname{Xi'an Jiaotong-Liverpool University}, \orgaddress{
\city{Suzhou}, \postcode{215123}, 
\country{China}}}

\affil*[4]{\orgdiv{Institute of Deep Perception Technology}, \orgname{JITRI}, \orgaddress{
\city{Wuxi}, \postcode{214000}, 
\country{China}}}

\affil[5]{\orgdiv{Department of Mathematical Sciences}, \orgname{University of Liverpool}, \orgaddress{
\city{Liverpool}, \postcode{L69 7ZX}, 
\country{United Kingdom}}}

\affil[6]{\orgdiv{Faculty of Information Engineering and Automation}, \orgname{Kunming University of Science and Technology}, \orgaddress{
\city{Kunming}, \postcode{650500}, 
\country{China}}}


\abstract{
Natural language (NL) based vehicle retrieval is a task aiming to retrieve a vehicle that is most consistent with a given NL query from among all candidate vehicles. Because NL query can be easily obtained, such a task has a promising prospect in building an interactive intelligent traffic system (ITS). Current solutions mainly focus on extracting both text and image features and mapping them to the same latent space to compare the similarity. However, existing methods usually use dependency analysis or semantic role-labelling techniques to find keywords related to vehicle attributes. These techniques may require a lot of pre-processing and post-processing work, and also suffer from extracting the wrong keyword when the NL query is complex. To tackle these problems and simplify, we borrow the idea from named entity recognition (NER) and construct FindVehicle, a NER dataset in the traffic domain. It has 42.3k labelled NL descriptions of vehicle tracks, containing information such as the location, orientation, type and colour of the vehicle. FindVehicle also adopts both overlapping entities and fine-grained entities to meet further requirements. To verify its effectiveness, we propose a baseline NL-based vehicle retrieval model called VehicleFinder. Our experiment shows that by using text encoders pre-trained by FindVehicle, VehicleFinder achieves 87.7\% precision and 89.4\% recall when retrieving a target vehicle by text command on our homemade dataset based on UA-DETRAC \cite{wen2020ua}. From loading the command into VehicleFinder to identifying whether the target vehicle is consistent with the command, the time cost is 279.35 ms on one ARM v8.2 CPU and 93.72 ms on one RTX A4000 GPU, which is much faster than the Transformer-based system. The dataset is open-source via the link https://github.com/GuanRunwei/FindVehicle, and the implementation can be found via the link https://github.com/GuanRunwei/VehicleFinder-CTIM.

}

\keywords{Cross modal learning, Named entity recognition, Intelligent traffic system, Vehicle retrieval, Human-computer interaction, Object detection }



\maketitle

\section{Introduction}

Vehicle retrieval is a task that aims to find the target vehicle from a large image gallery given a query image, which is an image-to-image matching technique also known as vehicle re-identification \cite{LiuHongye2016DeepRD}\cite{XinchenLiu2016ADL}\cite{XinchenLiu2016LargescaleVR}\cite{adaimi2021deep}. It has promising prospects in building ITS \cite{el2020pedestrian}\cite{ganin2019resilience}\cite{chien2020low}\cite{sharma2022vehicle} for smart cities \cite{kong2020multimedia}. However, an image-based vehicle retrieval system also has its defects in practice. For example, such a system needs an image to provide characteristics of the target vehicle, which is not always easy to obtain in the real world. The performance of an image-based vehicle retrieval system may also be limited because there is only one type of modality to provide spatial and temporal information.

To alleviate these problems, Natural Language (NL), as another essential modality in the real world, has received more and more attention from researchers in recent years. A natural language-based vehicle retrieval system aims to identify the target vehicle using an NL description. Such a system has a broader range of application scenarios, such as finding a vehicle when a bystander provides only an informal description. Most current natural language vehicle retrieval implementations construct the text encoder and visual encoder to extract features from both data types. They then project the obtained text and visual embeddings into the same latent space to compare their similarity. In addition, both visual and NL data will be carefully modified by these methods for more effective representation.  For example, vehicle track images are cropped to generate a global motion image \cite{park2021keyword}\cite{ChuyangZhao2022SymmetricNW}\cite{bai2021connecting}\cite{BochengXu2022NaturalLV}. As for NL, some keywords related to vehicle attributes (e.g., colour, vehicle type and orientation) are extracted in the given NL query \cite{park2021keyword}\cite{ChuyangZhao2022SymmetricNW}\cite{BochengXu2022NaturalLV}\cite{nguyen2021contrastive}. Although these works achieve acceptable performance on the CityFlow-NL \cite{QiFeng2021CityFlowNLTA} benchmark, they can still be improved, especially in terms of NL. We find that when implementing keyword extraction, existing methods are usually based on dependency analysis (e.g., using NLTK package) or semantic role labelling techniques to determine whether the word is a keyword or not. These techniques only assign the part of speech to the words in the sentence. It means that pre-determined rules and post-processing are required to divide the extracted keywords into corresponding vehicle attributes, making the whole process complex \cite{nguyen2021contrastive}\cite{JiachengZhang2022AMR}. Such methods can also suffer from extracting the wrong keyword if the NL description is complex. This can lead to error propagation in subsequent modules and reduce model performance.

In fact, keyword extraction is already a mature technology in natural language processing (NLP), also known as named entity recognition (NER). The main obstacle that prevents us from applying the state-of-the-art NER model to solve the above problem is the lack of a domain-specific corpus with high-quality annotations. Therefore, to alleviate this problem, we propose a named entity labelled natural language dataset focused on the traffic domain, called FindVehicle. It consists of descriptions of the vehicle from the point of view of urban traffic surveillance cameras. Some example descriptions from our dataset are shown. We also compare them with instances selected from other traffic domain datasets using natural language, namely Talk2Car \cite{deruyttere2019talk2car} and CityFlow-NL \cite{QiFeng2021CityFlowNLTA}. All details are given in Table \ref{traffic_nlp}. We carefully construct the vehicle descriptions to match real traffic scenarios and to enrich more detailed information about the target vehicles. Our dataset includes eight types of vehicle features, namely vehicle location, orientation, brand, model, type, colour, distance from the traffic surveillance camera, and velocity.
In contrast, Talk2Car \cite{deruyttere2019talk2car} only records vehicle type, and CityFlow-NL \cite{QiFeng2021CityFlowNLTA} has only four types of information, which are vehicle colour, type, action, and scene. More vehicle information in the description text means that the data can more accurately reflect the traffic scene in real life while reducing the challenge in NL-based vehicle retrieval tasks caused by the ambiguity of natural language. Both FindVehicle and CityFlow-NL \cite{QiFeng2021CityFlowNLTA} have the description of the relationship with other vehicles (surrounding vehicles). Therefore, we do not treat the surrounding vehicle as a separate feature. Furthermore, FindVehicle is annotated with multi-granularity named entity labels in order to be able to meet further requirements in the future.

To verify the effectiveness of the proposed dataset, we construct a simple and highly efficient cross-modal vehicle retrieval system called VehicleFinder. Unlike current transformer-based models \citep{radford2021learning}\cite{rao2022denseclip}, which have huge parameters and slow inference time, VehicleFinder has only 8.81 million parameters. This means that it can achieve real-time performance in the actual scenario and is more friendly to edge devices. VehicleFinder is trained and tested on our homemade text-to-image dataset called Vehicle-TI based on the training set of UA-DETRAC \cite{wen2020ua}. The keywords fed into VehicleFinder are extracted by a NER model pre-trained on FindVehicle. The experiment result shows that VehicleFinder gets 87.7\% precision and 89.4\% recall when detecting the vehicle. Its latency is 279.35 ms on one 8-core ARM v8.2 CPU.

To conclude, the main contributions of this paper include:

\begin{enumerate}
    \item We propose the first NER dataset (benchmark) in the traffic domain called FindVehicle, which has 42.3 thousand sentences, 1.361 million tokens, 202.5 thousand entities and 21 entity types. FindVehicle is not only a dataset that contains both flat and overlapping entities, but also has both coarse-grained and fine-grained entity types.
    \item We propose a text-image cross-modal vehicle retrieval system called VehicleFinder to prove the effectiveness of our proposed NER dataset. VehicleFinder is a highly efficient model with favourable performance that can achieve real-time performance and be applied to edge devices.
    \item During the experiment, we construct a text-image vehicle matching dataset called Vehicle-TI. Vehicle-TI has 335,040 training samples, 179,520 test samples and 83,776 validation samples.
\end{enumerate}

The rest of this paper is organized as follows: Section \ref{related_work} presents the related work of our paper; Section \ref{construction_findvehicle} presents the critical information of FindVehicle and how we construct it; Section \ref{statistics_findvehicle} presents the statistics details of FindVehicle; Section \ref{section_vehiclefinder} presents VehicleFinder, our text-image cross-modal vehicle retrieval system; Section \ref{experiments_findvehicle} includes the baselines of FindVehicle; Section \ref{experiments_VehicleFinder} presents the experiment details of VehicleFinder; Section \ref{conclusion_future} presents the conclusion of this paper and our future work; Section \ref{discussion} presents some challenges of FindVehicle.

\begin{table*}
\tiny
\setlength\tabcolsep{1pt}
\caption{Datasets of Vehicle Retrieval}
\centering
\label{traffic_nlp}
\begin{tabular}{lllll} 
\toprule   
  Dataset & Data Samples & Informative$[1]$ & Amount & HasNER \\  
\midrule   
    & My friend is getting out of the car. Stop and let me out too! & &  &  \\
   Talk2Car$[2]$ & Yeah that would be my son on the stairs next to the bus. & Low & 11959 & No \\
    & My mum is on the right! Park near her, she might want a lift.  & &  \\
\midrule 
    & A cargo truck drives down an intersection with many smaller cars. &  &  \\
   CityFlow-NL$[3]$ & The large green flatbed 18 wheeler is going straight. & Medium & 9374 & No \\
    & A green truck drives through an intersection, followed by a sedan. &  &  \\
\midrule
  & A [grey] [sedan] drives [right] at a speed of [58km per hour]. &  &  \\
   & Maybe a [[Ford] [Mondeo]]$[4]$. &  &  \\
   FindVehicle & A [Volvo] [truck] is parked on the side of the road , behind a  & High & 42341 & Yes \\
   &  [[Ferrari] [458]]. &  &  \\
   &  The [blue] [[BMW] [320]] driving [away] is [150 meters] away  &  &  \\
   & from us. & & \\
\bottomrule  
\end{tabular}
\\
\footnotesize{
$[1]$ Calculated based on the concentration of key vehicle features in the text description. \\
$[2]$ \citep{deruyttere2019talk2car} \\
$[3]$ \citep{QiFeng2021CityFlowNLTA} \\
$[4]$ It refers to a span that is allocated to one entity type. For example, in the phrase [[Ford] [Mondeo]], [Ford] is an \textit{vehicle\_brand} entity and [Mondeo] is a \textit{vehicle\_model} entity. [[Ford] [Mondeo]] is an entity of \textit{vehicle\_type-sedan}.
} \\
\end{table*}

\section{Related Work}
\label{related_work}

\subsection{Named Entity Recognition}
Named entity recognition (NER) is a classical sequence tagging task in NLP. It is to locate and classify the words or sentences with specific types in the text. The input of NER model is a sequence with part-of-speech (POS) taggings, as it shows in Equation \ref{ner_input},

\begin{equation}
WT = (w_1, t_1) ,  (w_2, t_2)  \dots  (w_i, t_i)  \dots  (w_n, t_n)
\label{ner_input}
\end{equation}

where $n$ denotes the number of words segmented by word segmentation program. $t_i$ is the POS of the word $w_i$.

The process of NER based on word segmentation and POS tagging is splitting, combining (determining entity boundaries) and reclassifying (determining entity categories) some words. The output is an optimal sequence $WC^*, TC^*$ with a pair format of \textit{(word category (WC), tagging category (TC))}, as it shows in Equation \ref{ner_output},

\begin{equation}
WC^*, TC^* = (wc_1, tc_1), (wc_2, tc_2), (wc_i, tc_i), \dots , (wc_m, tc_m)
\label{ner_output}
\end{equation}

where $m \leq n$, $wc_i = [w_j,\dots,w_{j+k}]$, $tc_i=[t_j, \dots, t_{j+k}]$, $1 \leq k$, $j + k \leq n$.

In brief, the NER modal could be written as Equation \ref{ner_model} shows,

\begin{equation}
(WC^*, TC^*) = argmax_{(WC, TC)}P(WC, TC \vert W, T)
\label{ner_model}
\end{equation}

where $W$ is the word sequence while $T$ is the tagging sequence. $P(\cdot)$ is a conditional probability model.

Hidden Markov Models \citep{morwal2012named} and Conditional Random Fields \citep{xu2008crf} are two typical machine learning models for NER. Convolutional neural network \citep{gui2019cnn}, recurrent neural network \citep{huang2015bidirectional}, transformer \citep{li2020flat}, and graph neural network \citep{sui2022trigger}, these deep learning models all achieve the state-of-the-art results in NER. 

Moreover, many NER datasets have been proposed in past years. These \cite{sang2003introduction}\cite{balasuriya2009named}\cite{derczynski2017results}\cite{weischedel2013ontonotes}\cite{ding2021few} are the well-known NER datasets (benchmarks). In these datasets, there are mainly three kinds of named entities: flat entity, overlapped entity and discontinuous entity. \cite{li2021unified} proposed a unified neural framework to concurrently solve the three NER problems.

\subsection{Text-Image Vehicle Retrieval}
Vehicle retrieval based on test-image cross-modal learning is a hot spot these years \cite{park2021keyword}\cite{scribano2021all}\cite{bai2021connecting}\cite{khorramshahi2021towards}\cite{sun2021dun}\cite{BochengXu2022NaturalLV}\cite{le2022tracked}\cite{nguyen2021contrastive}\cite{tt2021deep}. The model could find out the highest matching vehicle based on the description with the text format. There are mainly two formats according to the architecture. The first is the end-to-end neural network based on early retrieval, where the features of images and text are fused in the early stage. The second is the non-end-to-end system based on late retrieval, where images and text features are extracted individually and loaded into a decision module. 

\subsection{Contrastive Language Image Pretraining}
Contrastive language image pretraining (CLIP) combines the modalities of language and image in one neural network, which is mainly for multi-modal tasks based on natural language and computer vision. Prior to this, most computer vision work was trained based on pre-defined labels, and supervision limited the generalization and usefulness of neural networks. There has been a lot of work in the field of NLP using a large amount of corpus data for self-supervised learning, and the effect of these models has surpassed manually labelled datasets \cite{devlin2018bert}\cite{floridi2020gpt}. In the field of CV, the current mainstream method is still to use large-scale datasets with labelled information for pre-training \citep{deng2009imagenet}. Vanilla CLIP \citep{radford2021learning} creatively uses text as a supervision signal to train a vision model and achieves conspicuous results on ImageNet \citep{deng2009imagenet}. In addition, vanilla CLIP \citep{radford2021learning} is also very good at zero-shot tasks. \cite{rao2022denseclip} proposes a CLIP framework called DenseCLIP, which is good at dense prediction tasks, such as semantic segmentation and dense object detection. \cite{goel2022cyclip} proposes a new contrastive loss to normalize the location and geometric information of image and text features in the semantic space.

\section{The Construction of FindVehicle}
\label{construction_findvehicle}

\subsection{Brief Introduction}
FindVehicle is the first NER dataset in traffic. It is based on the image samples of UA-DETRAC \citep{wen2020ua}. FindVehicle contains various descriptions of traffic participants on the road from the view of traffic surveillance cameras, mainly vehicles. A description contains many attributes of one or several vehicles. These attributes all could be detected by traffic sensors, such as surveillance cameras, lidar and radar. 
Moreover, FindVehicle also incorporates much real-world prior knowledge, such as the vehicle brand and model. Furthermore, FindVehicle contains both coarse-grained and fine-grained entities. Entities include both flat and overlapped entities.

\begin{figure*}
\centerline{\includegraphics[width=.93\textwidth]{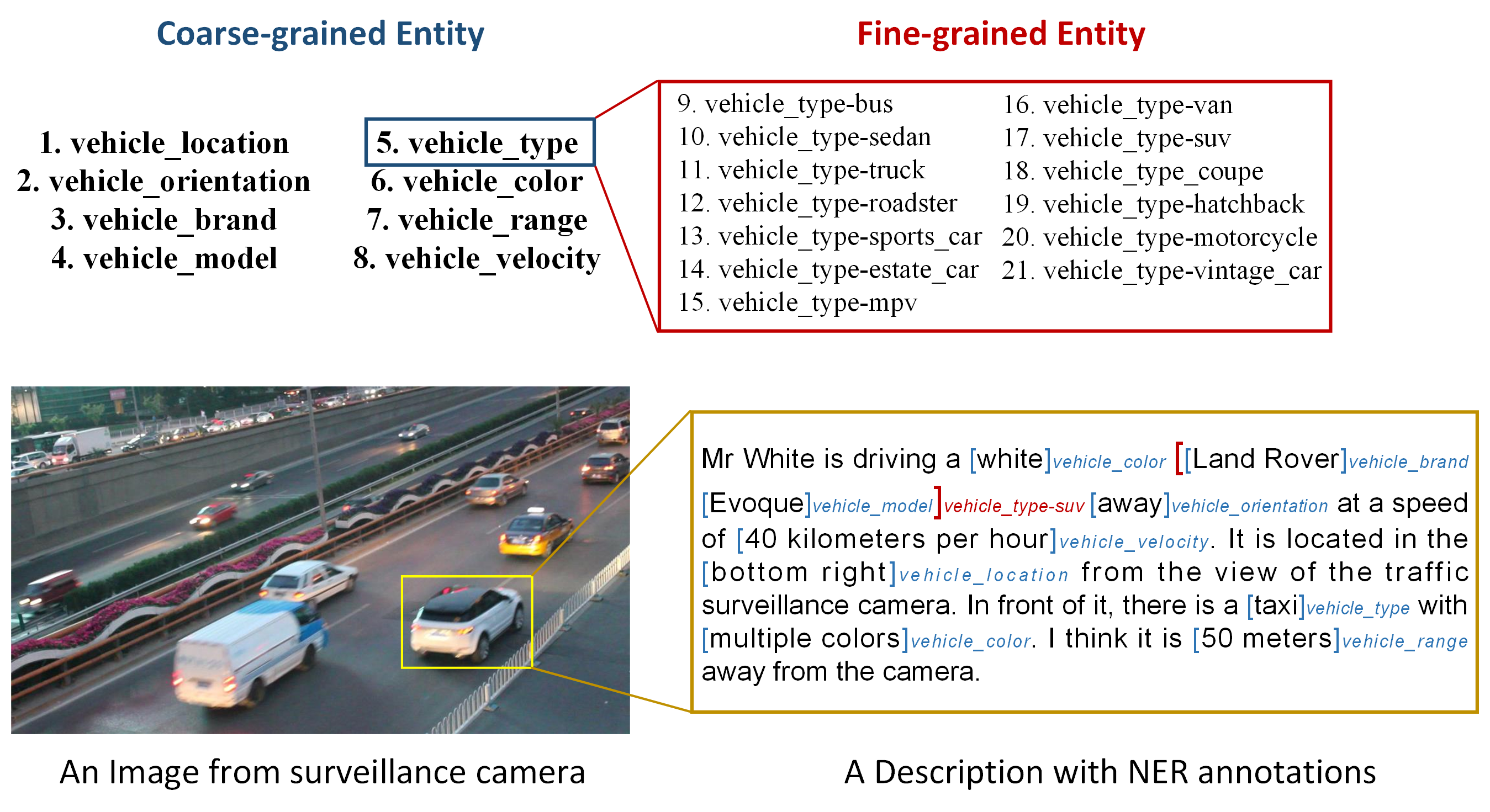}}
\caption{Entity types and an annotated sample of FindVehicle. Images are from UA-DETRAC \citep{wen2020ua}.}
\label{ner_entity}
\end{figure*}

\subsection{Entity Types}

As Fig. \ref{ner_entity} shows, there are 21 entity types in FindVehicle, 8 coarse-grained entities and 13 fine-grained entities. These entities are all the attributes of vehicles, which all follow the distribution of the real world. Moreover, FindVehicle also contains both flat and overlapped entities.

\subsubsection{Coarse-grained Entity}

There are 8 kinds of coarse-grained entities, including \textit{vehicle\_location}, \textit{vehicle\_orientation}, \textit{vehicle\_brand}, \textit{vehicle\_model}, \textit{vehicle\_type}, \textit{vehicle\_color}, \textit{vehicle\_range} and \textit{vehicle\_velocity}.

\textit{vehicle\_location} indicates the locations of vehicles from the view of the traffic surveillance cameras, such as \verb|bottom right|, \verb|top-left|, etc.

\textit{vehicle\_orientation} indicates the directions of vehicles' heads from the view of the traffic surveillance cameras, such as \verb|this way|, \verb|away|, etc.

\textit{vehicle\_brand} indicates the brands of vehicles. FindVehicle contains 65 vehicle brands all over the world.

\textit{vehicle\_model} indicates the models of vehicle brands. There are 4793 models of different vehicle brands in FindVehicle. For example, \verb|Q7| is one of the models of \verb|Audi|. 

\textit{vehicle\_type} indicates the types of vehicles, such as \verb|sedan|, \verb|suv|, etc.

\textit{vehicle\_color} indicates the colors of vehicles, such as \verb|silver grey|, \verb|rose red|, etc.

\textit{vehicle\_range} indicates the distance between the vehicle and the traffic surveillance camera, such as \verb|18m|, \verb|123 meters|, etc.

\textit{vehicle\_velocity} indicates the speed of the moving vehicle on the road, such as \verb|50 kilometres per hour|, \verb|120 km/h|, etc.

\subsubsection{Fine-grained Entity}

As it shows in Fig. \ref{ner_entity}, in FindVehicle, there are 13 kinds of fine-grained entities, which belong to the coarse-grained entity \textit{vehicle\_type}, for example, \verb|BMW X5| is a fine-grained entity of \textit{vehicle\_type-suv}. Fine-grained entities contain the human prior knowledge of cars. 

\subsubsection{Flat and Overlapped Entity}
\begin{figure}
\centerline{\includegraphics[width=.35\textwidth]{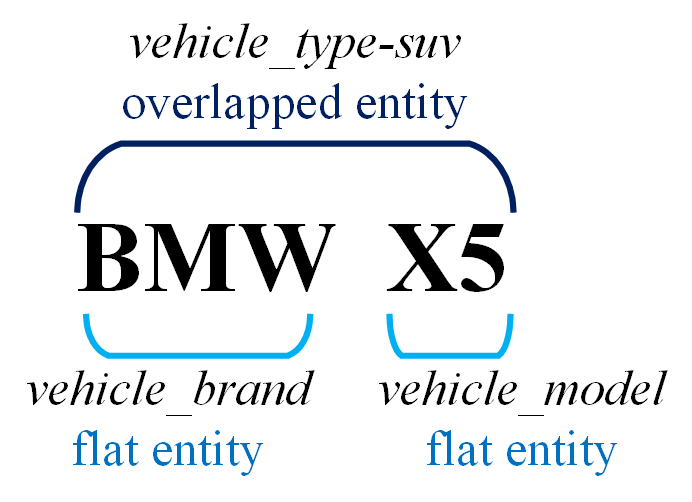}}
\caption{An example of flat and overlapped entities.}
\label{overlapped_entity}
\end{figure}

Overlapped entities exist in coarse-grained entities \textit{vehicle\_brand}, \textit{vehicle\_model} and fine-grained entities \textit{vehicle\_type-*}. For example, as Fig. \ref{overlapped_entity} shows, the label of \verb|BMW| is \textit{vehicle\_brand} while the label of \verb|X5| is \textit{vehicle\_model}, for a car enthusiast, the label of \verb|BMW X5| is \textit{vehicle\_type-suv}.

\subsection{Corpus Collection}

\begin{figure}
\centerline{\includegraphics[width=.65\textwidth]{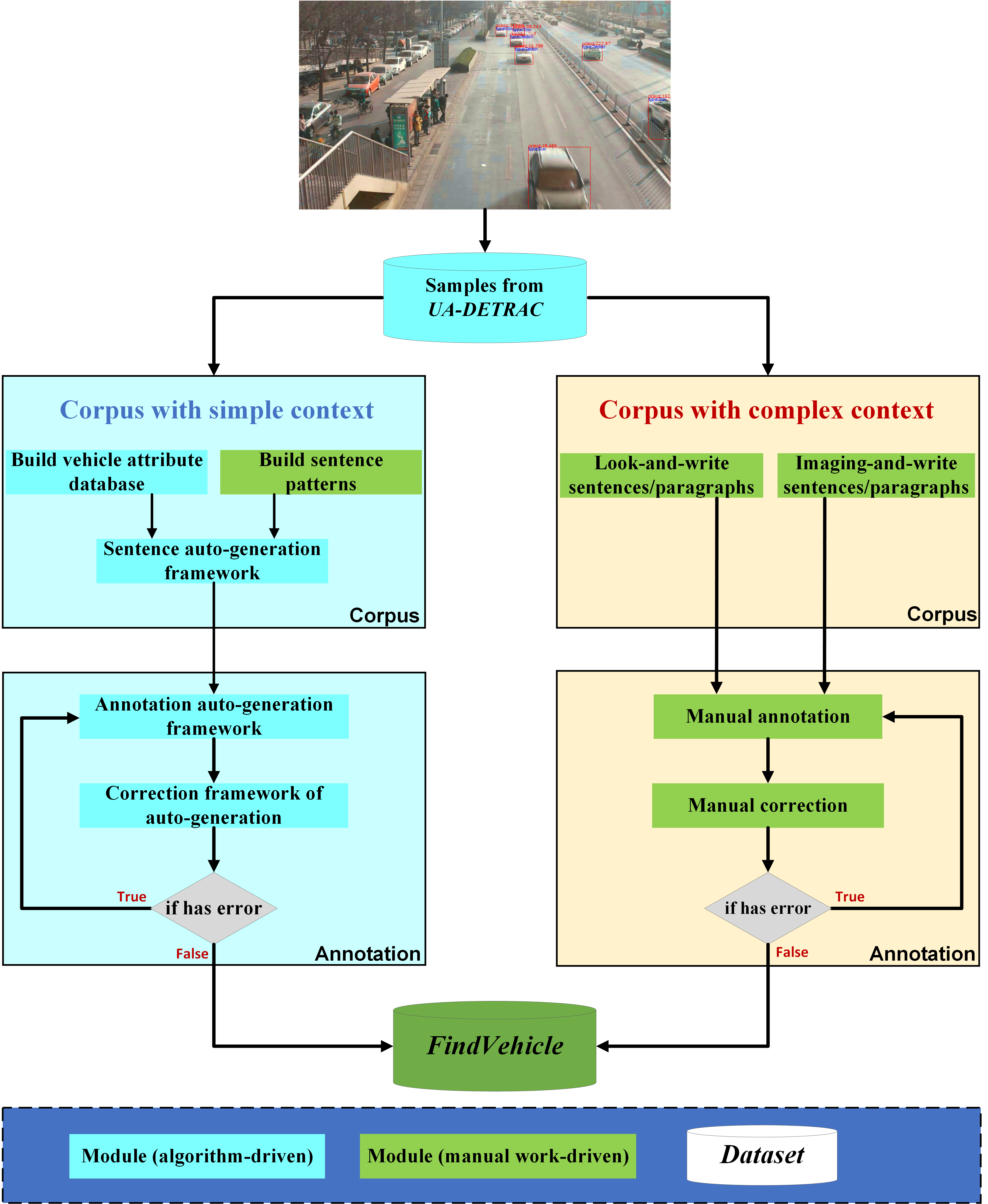}}
\caption{The framework of corpus collection and annotation of FindVehicle.}
\label{ner_generation}
\end{figure}

As Fig. \ref{ner_generation} shows, the corpus collection includes two parts, the corpus with simple context and the corpus with complex context. 

\begin{figure}
\centerline{\includegraphics[width=.89\textwidth]{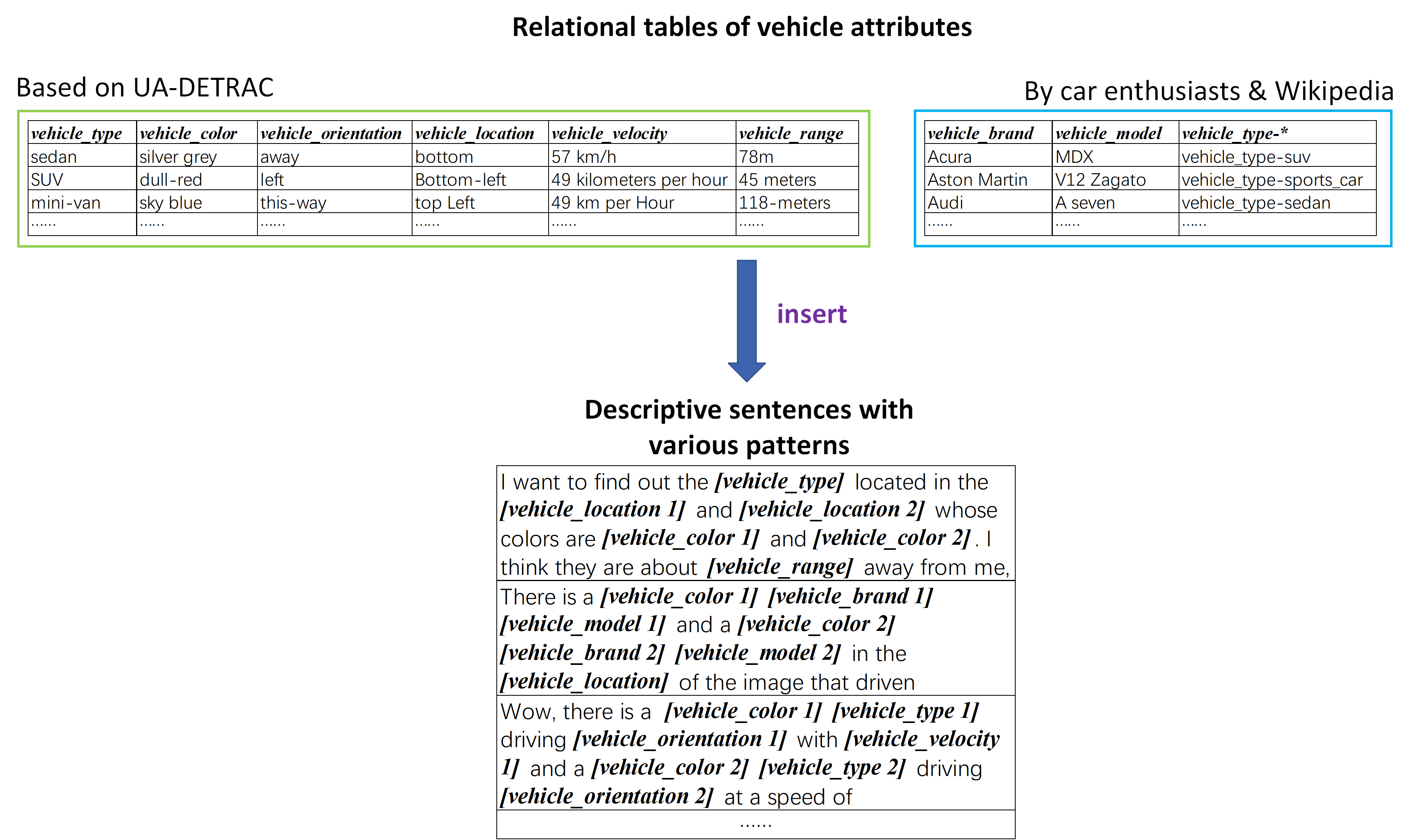}}
\caption{The generation of corpus with simple context.}
\label{simple_sample_framework}
\end{figure}

The corpus with simple context denotes the short sentences, which are presented in the column of Data Samples in Table \ref{traffic_nlp}. As Fig. \ref{simple_sample_framework} presents, firstly, we sample some target vehicles with bounding boxes and labels in UA-DETRAC \citep{wen2020ua}. Based on these samples, we create a relational table to save the attributes of the corresponding vehicle. Each item in the table represents one vehicle with several attributes. Furthermore, to increase the complexity of the dataset, we replace some formal phrase-type and word-type entities with our informal expression habits and add some rare entities which do not exist in UA-DETRAC \citep{wen2020ua}. Moreover, for the entity generation of three entities \textit{vehicle\_brand}, \textit{vehicle\_model} and \textit{vehicle\_type-*}, we invite three car enthusiasts to collect and integrate data based on their extensive car knowledge and the search results of Wikipedia. They write data with different expressions and curate 65 vehicle brands, 4793 vehicle models and 13 vehicle types in total. Secondly, we recruit four volunteers to write descriptive sentences with various patterns in their tone and expression habits. All volunteers are well-educated and have adequate English linguistic knowledge. Thirdly, we insert the target vehicles with their attributes into these patterns by our sentence auto-generation framework. 

As the sample in Fig. \ref{ner_entity} presents, the corpus with complex context indicates narrative long sentences or paragraphs with persons’ subjective emotions and imagination. Instead of generating a corpus with simple context by combining labor and computers, a corpus with complex context is made by human beings only. Four members of our team write down the corresponding sentences and paragraphs with their own writing habits and imagination by observing the images in UA-DETRAC \citep{wen2020ua}.

\begin{figure}
\centerline{\includegraphics[width=.65\textwidth]{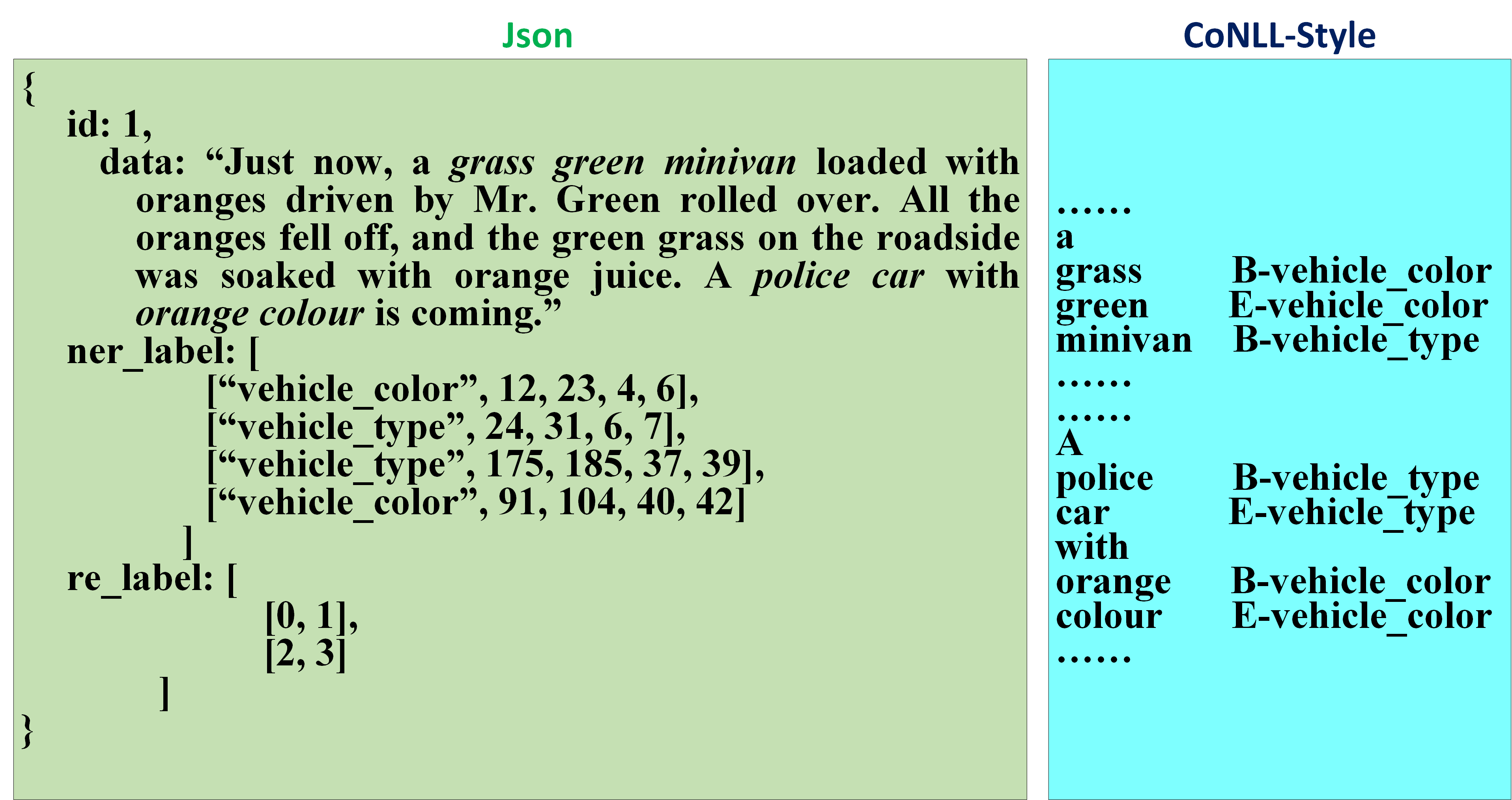}}
\caption{The two annotation formats of FindVehicle.}
\label{ner_annotation_format}
\end{figure}

\subsection{NER Annotation}

As Fig. \ref{ner_generation} shows, in our NER annotation framework, there are two processes for the corpus with simple and complex contexts, respectively.

The annotations of the corpus with simple context are completed simultaneously with sentence auto-generation by our annotation auto-generation framework. After that, the correction framework of auto-generation will automatically identify whether the NER annotations by the auto-generation framework have errors. If the data had an error, the annotation process would be interrupted and report the location of the error, and then we would check and fix it. If it had no error, the corpus with annotations would be loaded into the dataset directly. 

The annotations of the corpus with complex context are totally manual. They are based on the common sense and knowledge of annotators. Annotators are all volunteers who are knowledgeable about vehicles and good at narrative writing.

As Fig. \ref{ner_annotation_format} shows, we organize the data in two formats, JSON and CoNLL-style \citep{sang2003introduction}. The value of the key \textit{ner\_label} is the annotated named entities. The values of \textit{ner\_label} in each element is \textit{[entity type, start index of char span, end index of char span, start index of token span, end index of token span]}. Our annotation considers char-level and token-level, meeting different needs of the NER models. The key \textit{re\_label} denotes the indexes of values of \textit{ner\_label} that refer to one target in the context of a sentence.

\section{Data Statistics of FindVehicle}
\label{statistics_findvehicle}

\begin{table}
\caption{Statistics of FindVehicle and other well-known NER datasets}
\centering
\label{ner_datasets}
\begin{tabular}{llllll}  
\toprule   
  Datasets & Sentences & Tokens & Entities & Entity Classes & Domain \\  
\midrule   
WikiGold \citep{balasuriya2009named} & 1.7k & 39k & 3.6k & 4 & General \\
WNUT'17 \citep{derczynski2017results} & 4.7k & 86.1k & 3.1k & 6 & Social Media \\
CoNLL'03 \citep{sang2003introduction} & 22.1k & 301.4k & 35.1k & 4 & Newswire \\
I2B2 \citep{stubbs2015annotating} & 107.9k & 805.1k & 28.9k & 23 & Medical \\
OntoNotes \citep{weischedel2013ontonotes} & 103.8k & 2067k & 161.8k & 18 & General \\
\midrule
\textbf{FindVehicle (ours)} & \textbf{42.3k} & \textbf{1361.1k} & \textbf{202.5k} & \textbf{} & \textbf{Traffic} \\
\bottomrule  
\end{tabular}
\end{table}

\subsection{Size and Distribution of FindVehicle}

\begin{figure*}
\centerline{\includegraphics[width=.96\textwidth]{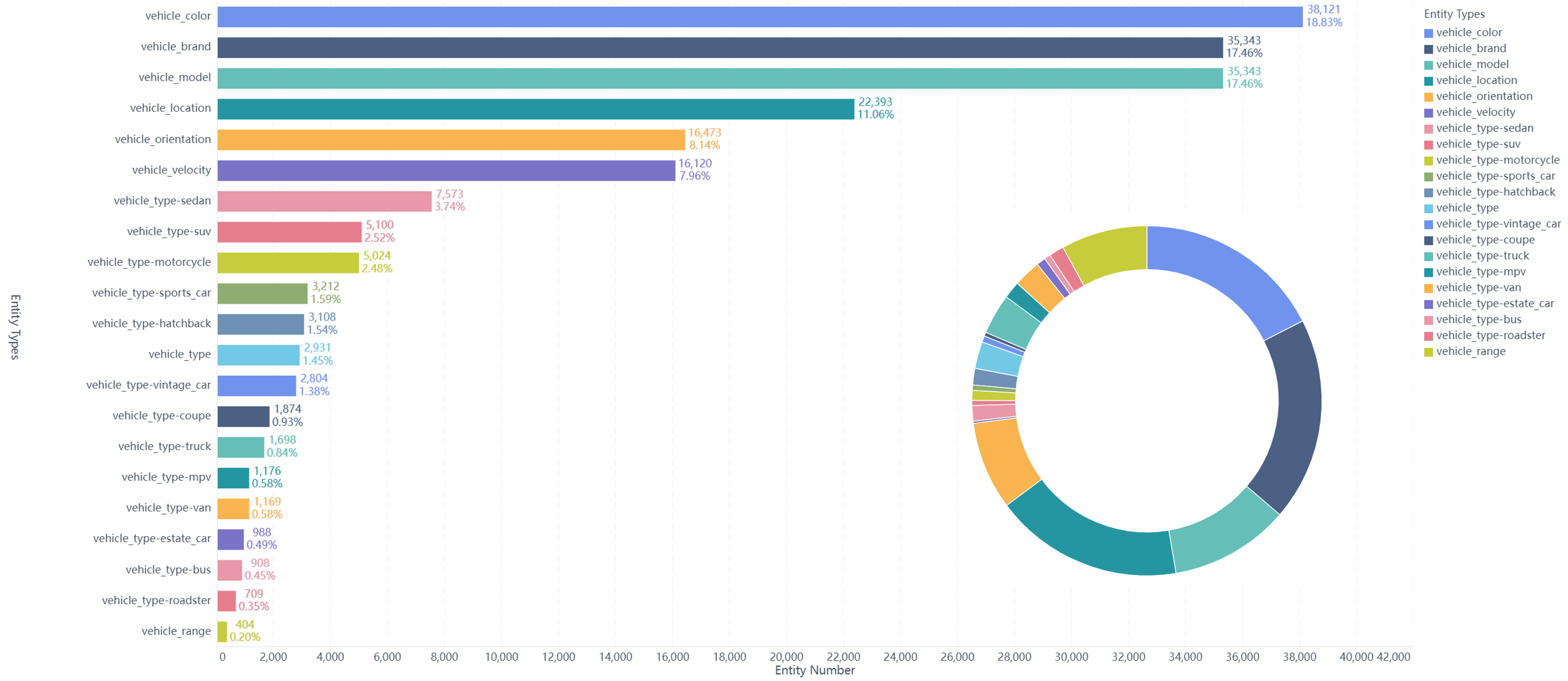}}
\caption{Statistics by entities in FindVehicle.}
\label{ner_entity_statistics}
\end{figure*}

FindVehicle is the first NER dataset in traffic with the annotations of automatic labeling and
manual labelling together. As Table \ref{ner_datasets} shows, we present the statistics of FindVehicle and other widely used well-known NER datasets, including CoNLL'03 \citep{sang2003introduction}, WikiGold \citep{balasuriya2009named}, WNUT'17 \citep{derczynski2017results}, I2B2 \citep{stubbs2015annotating} and OntoNotes \citep{weischedel2013ontonotes}. FindVehicle has 42.3 thousand sentences, 1.361 million tokens, 202.5 thousand entities and 21 entity classes. As Fig. \ref{ner_entity_statistics} presents, the entity types are long-tail distributed to reflect the real-world traffic scenario.

\subsection{Dataset Split}

FindVehicle is a hybrid NER dataset containing both flat and overlapped entities. We split it into a training set and a test set. The details of these two sets are shown in Table \ref{ner_datasets_split}. For the training set, there are 84.6k coarse-grained entities and 18.2k fine-grained entities. In addition, there are 84.2k flat entities and 18.6k overlapped entities. For the test set, there are 82.5k coarse-grained entities and 17.4k fine-grained entities. Besides, 82.7k flat entities and 17.2k overlapped entities are in the test set.

\begin{table}
\tiny
\caption{Data Split of FindVehicle}
\centering
\label{ner_datasets_split}
\begin{tabular}{lllll}  
\toprule   
Dataset & Coarse-grained Entities & Fine-grained Entities & Flat Entities & Overlapped Entities \\
\midrule   
Training set & 84.6k & 18.2k & 84.2k & 18.6k \\
Test set & 82.5k & 17.4k & 82.7k & 17.2k \\
\bottomrule  
\end{tabular}
\end{table}

\begin{figure*}
\centerline{\includegraphics[width=.96\textwidth]{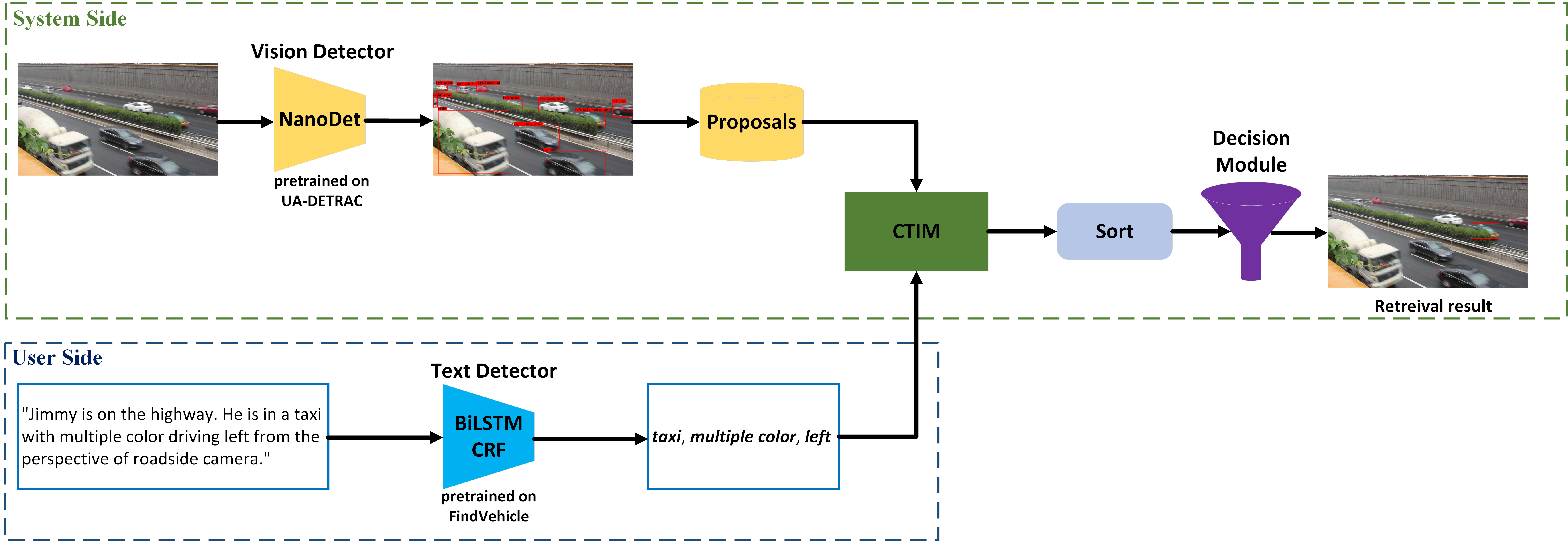}}
\caption{The architecture of VehicleFinder.}
\label{VehicleFinder}
\end{figure*}

\section{VehicleFinder}
\label{section_vehiclefinder}

VehicleFinder is a lightweight text-image cross-modal vehicle retrieval system. Users could find out the target vehicle through the description of its type, color and orientation. As Fig. \ref{VehicleFinder} presents, VehicleFinder has two branches. One is to extract proposals by a vision detector while the other is to extract named entities by a text detector. We adopt NanoDet \citep{nanodet} as the vision detector and BiLSTM-CRF \citep{huang2015bidirectional} as the text detector. The NanoDet \citep{nanodet} is pretrained on UA-DETRAC \citep{wen2020ua} while the BiLSTM-CRF \citep{huang2015bidirectional} is pretrained on our FindVehicle. The proposals and name entities will be loaded into the contrastive text-image module (CTIM) to compare the semantic similarity of data of two modalities.

\begin{figure*}
\centerline{\includegraphics[width=.82\textwidth]{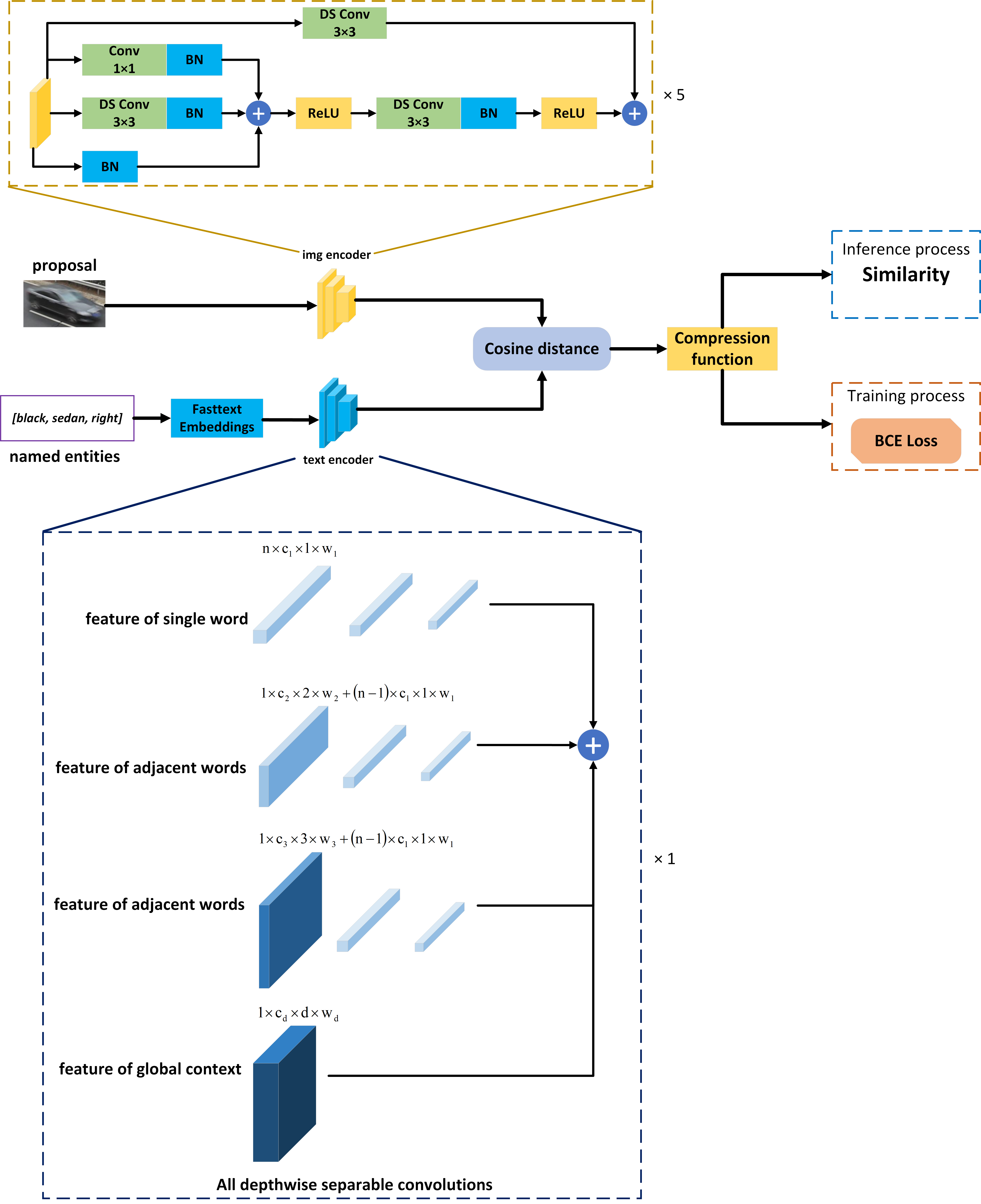}}
\caption{The architecture of Contrastive Text-Image Module (CTIM). All convolution operations are all depthwise separable convolutions, except for the convolution operation with the kernel size of $1 \times 1$. Because the depthwise separable convolution contains the convolution operation with the $1 \times 1$ kernel size. $c_i$ denotes the channel number of the feature map whose kernel height is $i$. $w_i$ denotes the width of the feature map whose kernel height is $i$.}
\label{CTIM}
\end{figure*}

As Fig. \ref{CTIM} shows, there are two encoder branches in CTIM to encode the data of image and text modalities, respectively. The output of CTIM is the similarity of the image and text, whose value domain is between 0 and 1. An output below 0.5 indicates that the image and text are unrelated, while an output above 0.5 indicates that they are related. CTIM is a complete convolution module whose convolution operations are all the depthwise separable convolution \cite{chollet2017xception}, dramatically reducing the parameter number, especially in the deep layers of the neural network. CTIM could perform as a plug-and-play module in some cross-modal systems.

In the branch of the image encoder, there are five same encoder units. An encoder unit will initially put the input feature map $x_i \in R^{c \times h \times w}$ into three branches, where $c, h, w$ respectively denote the channel, height and width of a feature map. The first three branches with different convolution kernel sizes are used to extract the feature with different receptive fields. The output feature map $\hat{x}_i \in R^{c \times h \times w} $ will be activated by ReLU \citep{agarap2018deep}, then increase the channels and reduce the spatial size through a depthwise separable convolution operation with the $3 \times 3$ kernel. After a batch normalization and a ReLU activation, the output feature map is $x_{i+1} \in R^{2c \times \frac{h}{2} \times \frac{w}{2}}$. To alleviate the gradient vanishing and explosion in the training stage, a long residual path with a depthwise separable convolution is connected with the output feature map. The final output of the encoder unit is $\hat{x}_{i+1} \in R^{2c \times \frac{h}{2} \times \frac{w}{2}}$. The whole process is presented in Equation \ref{img_encoder}.

\begin{align}
& \hat{x}_i = BN(Conv_{3\times 3}(x_i)) + BN(Conv_{1\times 1}(x_i)) + BN(x_i), \hat{x}_i \in R^{c \times h \times w} \nonumber \\
& x_{i+1} = BN(Conv_{3\times 3}(ReLU(\hat{x}_i))), x_{i+1} \in R^{2c \times \frac{h}{2} \times \frac{w}{2}} \nonumber \\
& \hat{x}_{i+1} = ReLU(x_{i+1}) + Conv_{3\times 3}(x_i), \hat{x}_{i+1} \in R^{2c \times \frac{h}{2} \times \frac{w}{2}}
\label{img_encoder}
\end{align}

\begin{figure}
\centerline{\includegraphics[width=.77\textwidth]{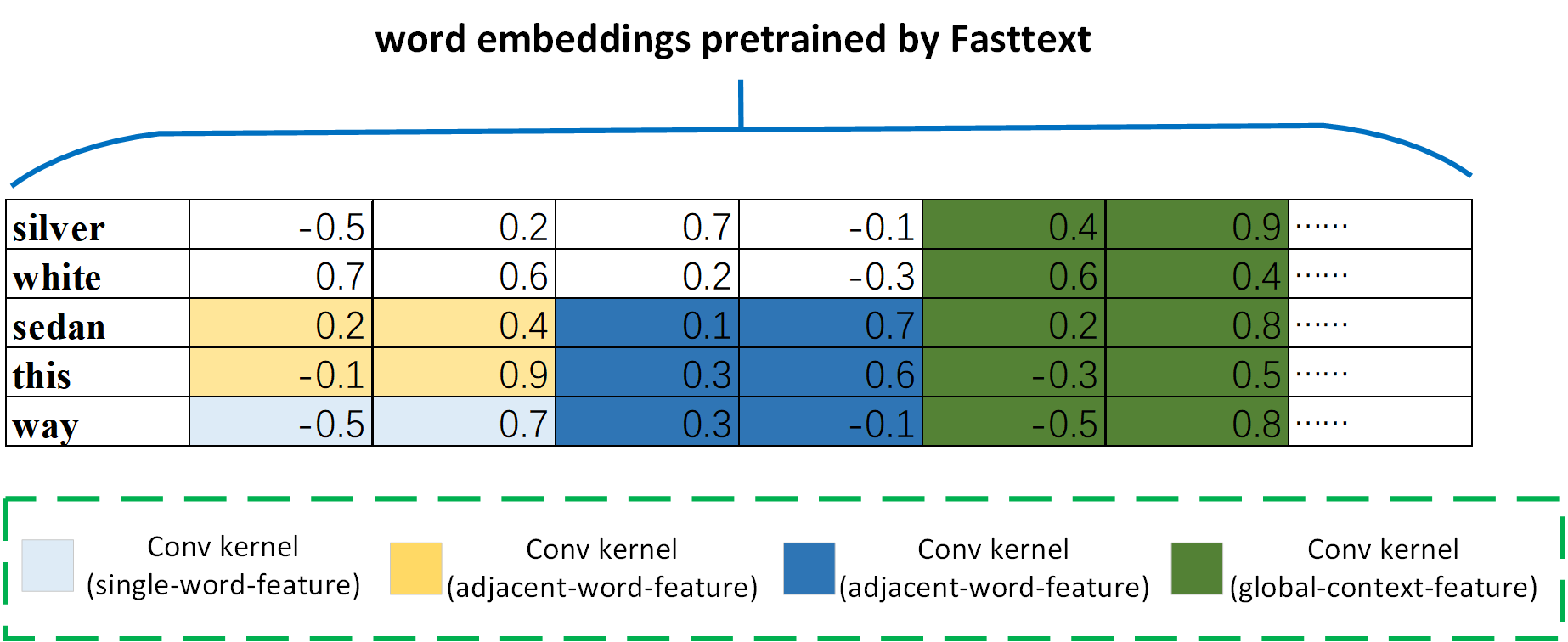}}
\caption{The four multi-scale convolution operations in our text encoder.}
\label{text_encoder_conv}
\end{figure}

In the branch of text encoder, named entities will be firstly embedded by pretrained embeddings of Fasttext (wiki-news-300d-1M) \citep{bojanowski2017enriching}. Fasttext could infer the embeddings of words not in the word dictionary based on the existing words, which is more robust than Word2vec \citep{mikolov2013efficient} and GloVe \citep{pennington2014glove} for the system. The shape of the embedding matrix is $d \times 300$, where $d$ indicates the number of named entities and 300 is the vector length of each named entity. After that, we adopt four groups of multi-scale depthwise separable convolution operations to extract the feature with different scales concurrently. The first group is $n$ convolution operations of the kernel size $1 \times w_1$, which is to extract the feature of a single word in named entities. The second group has one convolution operation of the kernel size $2 \times w_2$ and $n-1$ convolution operations of the kernel size $1 \times w_1$, where the convolution of the $2 \times w_2$ kernel is to extract the associated feature of adjacent words. The rest $1 \times w_1$ convolution operations are to enhance the non-linear representation. The third group is firstly processed by a convolution of the $3 \times w_3$ kernel, which is also for the feature extraction of adjacent words with a word window size of three. Then the following operations are the same as the second group. The fourth group is a convolution operation with the kernel size of $d \times w_d$, which is to extract the feature of the global context. Finally, the outputs of these four groups will be added to get a comprehensive representation of the name entities. The four convolution operations are shown in Fig. \ref{text_encoder_conv}.

After we get the representations of the proposal and named entities, we align their shape to calculate their cosine distance. Cosine distance measures the distance between vectors of the proposal and named entities. It could maintain the same similarity in the high-dimensional case as the low-dimensional case, which is a robust indicator of the relative difference in direction. Equation \ref{cosine_distance} shows cosine distance.

\begin{equation}
CosineD = \frac{\textbf{A} \cdot \textbf{B}}{\Vert \textbf{A} \Vert \Vert \textbf{B} \Vert} = \frac{\sum^n_{i=1}A_iB_i}{\sqrt{\sum^n_{i=1}A_i^2}\sqrt{\sum^n_{i=1}B_i^2}}, CosineD \in [-1, 1]
\label{cosine_distance}
\end{equation}

where $n$ is the number of vector's components. $A_i$ and $B_i$ respectively denote the text and image vector of $i$th component.

However, the value domain of cosine distance is $[-1, 1]$. It means that the result of cosine distance could not be directly fed to binary cross entropy loss (BCE loss) because BCE loss (Equation \ref{bce_loss}) could not process the negative number. 

\begin{equation}
    L_{BCE} = - \sum ^N_{i=1} [y_iln(\hat{y}_i) + (1-y_i)ln(1-\hat{y}_i)]
\label{bce_loss}
\end{equation}

where $N$ indicates the number of samples in a batch. $y_i \in \{0, 1\}$ is the ground truth while $\hat{y}_i \in [-1, 1]$ is the result of cosine distance predicted by the neural network. Apparently, $\hat{y}_i$ is not in the definitional domain of $ln(\cdot)$ if $\hat{y}_i$ is below zero.

Therefore, as Equation \ref{linear_compression} presents, we use Equation \ref{linear_compression} to compress the results of cosine distance from $[-1, 1]$ to $[0, 1]$, which can be the input to BCE loss. The linear compression function is a monotonically increasing function whose value domain is $[0, 1]$. It is differentiable everywhere. Monotonicity ensures that the relative position of the variable does not change when it maps from $[-1, 1]$ to $[0, 1]$. The property of being differentiated everywhere ensures that it can participate well in backpropagation in neural networks.

\begin{equation}
Comp(x) = \frac{1}{2}x + \frac{1}{2}, Comp(x) \in [0, 1]
\label{linear_compression}
\end{equation}

where $x \in [-1, 1]$ denotes the result calculated by cosine distance.

Therefore, the complete form of the loss function is presented in Equation \ref{final_loss}.

\begin{equation}
    L(y_i, \hat{y}_i) = - \sum ^N_{i=1} [y_iln(Comp(\hat{y}_i)) + (1-y_i)ln(1-Comp(\hat{y}_i))]
\label{final_loss}
\end{equation}

\begin{figure}
\centerline{\includegraphics[width=.72\textwidth]{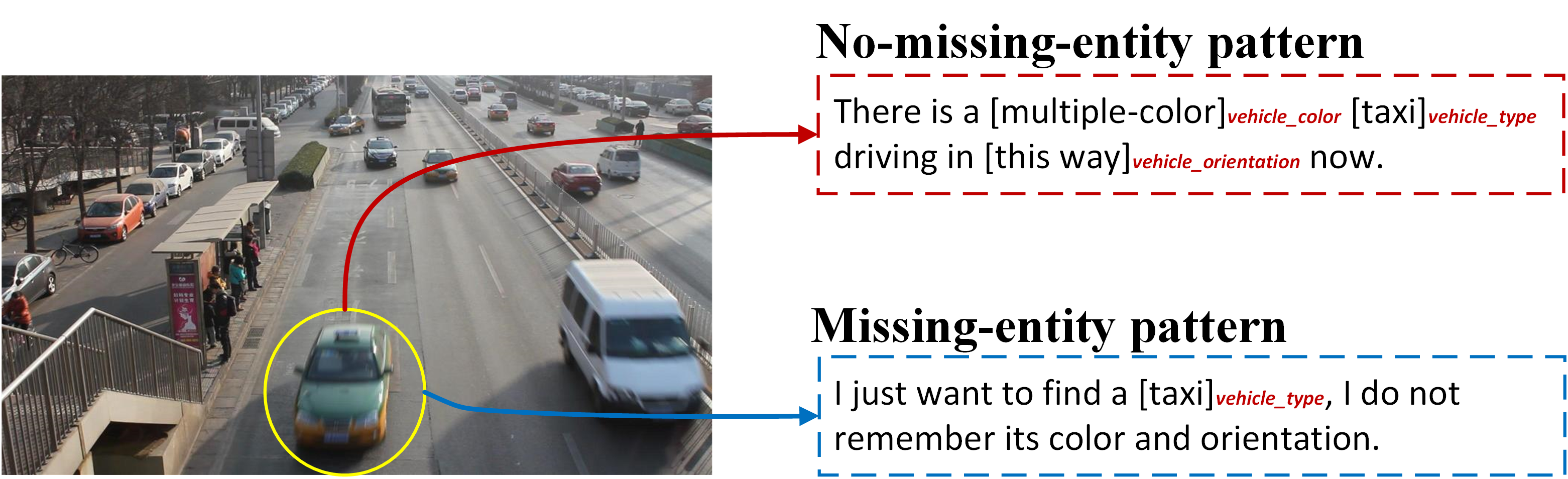}}
\caption{Two patterns of commands.}
\label{entity_patten_command}
\end{figure}

Finally, our VehicleFinder will calculate the similarities between named entities extracted from the command and object proposals extracted by the vision detector. We will sort object proposals in terms of the similarity with named entities descendingly. The ranking list $proposals$ will then be fed to a decision module. In the decision module, considering that the user cannot always describe the vehicle characteristics in detail, we take two patterns of commands into account and process them respectively to enhance the system's robustness, which could also be user-friendly. As Fig. \ref{entity_patten_command} presents, the first is the no-missing-entity pattern and the second is the missing-entity pattern. No-missing-entity pattern indicates the command contains all three named entities, \textit{vehicle\_type}, \textit{vehicle\_color} and \textit{vehicle\_orientation}. Missing-entity pattern indicates the command contains one or two named entities, and the other one or two named entities are not mentioned.

As Algorithm \ref{decision_module} presents, we firstly set two thresholds $th_{nm}$ and $th_{m}$, which mean the threshold for no-missing-entity pattern and the threshold for missing-entity pattern. The variable $proposals$ containing \textit{vehicle: sim} pairs indicates object proposals and their similarity with named entities extracted from the command.
For no-missing-entity pattern, if existing the \textit{vehicle: sim} pair whose $sim$ is larger than $th_{nm}$, the \textit{vehicle: sim} pair would be appended to $retainVehicle$. If not existing the \textit{vehicle: sim} pair whose $sim$ is larger than $th_{nm}$, the decision module would continue to search for the \textit{vehicle: sim} pair whose $sim$ is larger than $th_m$. If existing the \textit{vehicle: sim} pair whose $sim$ is larger than $th_m$, the \textit{vehicle: sim} pair would also be appended to $retainVehicle$. $th_{nm}$ and $th_{m}$ are set based on the results of experiments. We assume by default that $th_{nm}$ is greater than $th_m$.

\begin{algorithm}
\caption{Decision Module}\label{decision_module}
\begin{algorithmic}
\State \textbf{Input:} $th_{nm}$, $th_{m}$, $proposals=[\left\{vehicle:sim\right\}]$, $retainVehicle = [ ]$
\State \If{$proposals$.exists($vehicle.sim \geq th_{nm}$)} \\
       \quad \quad for $p$ in $proposals$ \textbf{do} \\
       \quad \quad \quad \quad \textbf{if} $p.sim \geq th_{nm}$ \textbf{do} \\
       \quad \quad \quad \quad \quad \quad $retainVehicle$.append($p$) \\
       \quad \quad \quad \quad \textbf{end if} 
\State \ElsIf{$proposals$.exists($th_{m} \leq vehicle.sim < th_{nm}$)} \\
       \quad \quad for $p$ in $proposals$ \textbf{do} \\
       \quad \quad \quad \quad \textbf{if} $p.sim \geq th_m$ \textbf{do} \\
       \quad \quad \quad \quad \quad \quad $retainVehicle$.append($p$) \\
       \quad \quad \quad \quad \textbf{end if} \\
       \quad \quad \textbf{end if} \\

       \EndIf
       
\State \textbf{return} $retainVehicle$
\end{algorithmic}
\end{algorithm}
\bigskip

\section{Experiments of FindVehicle}
\label{experiments_findvehicle}

In the experiments of FindVehicle, we make the baselines of our FindVehicle. 

\subsection{Settings of Training and Evaluation}
\label{training_setting_findvehicle}
We select three representative and state-of-the-art models to train and test on FindVehicle, which were BiLSTM-CRF \citep{huang2015bidirectional}, BERT-CRF \citep{souza2019portuguese} and FLERT \citep{schweter2020flert}. 

BiLSTM-CRF \citep{huang2015bidirectional} combines BiLSTM and CRF. BiLSTM acts as the encoder layer and takes word embeddings as input, CRF serves as a decoder to determine the tag for each token based on hidden states outputted from a encoder.

BERT-CRF \citep{souza2019portuguese} replaces word embeddings of BiLSTM with subword-embeddings learned from BERT, and changes the encoder from BiLSTM to Transformer.

FLERT \citep{schweter2020flert} is a NER model that takes document-level features as an extra account. By adding context text on both sides (left and right) to the query sentence, FLERT captures document-level features and presents a better predict result than the previous model.

For each model, we use the most suitable hyperparameters that make the model converge smoothly. We train and test these models on one TITAN RTX GPU. Table \ref{ner_model_hyper} shows the implementation details.

\begin{table}
\caption{Implementation Details of Models on The Training Set of FindVehicle}
\centering
\label{ner_model_hyper}
\begin{tabular}{llllll}  
\toprule   
  Model & Epochs & BS & ILR & Opt & Sch$^1$  \\
\midrule   
BiLSTM-CRF \citep{huang2015bidirectional} & 80 & 32 & 0.001 & AdamW & Cosine \\
BERT-CRF \citep{souza2019portuguese} & 100 & 4 & 0.0015 & AdamW & Cosine \\
FLERT \citep{schweter2020flert} & 80 & 8 & 0.001 &  AdamW & Cosine \\
\bottomrule  
\end{tabular}
\\
\footnotesize{$^1$ BS means batch size; ILR means initial learning rate; Opt means optimizer; Sch means scheduler.} 
\end{table}

Furthermore, as Equation \ref{precision}, \ref{recall} and \ref{f1} present, we choose precision, recall and F1 score as the evaluation metrics of the test, which are based on the confusion matrix (Table \ref{confusion_matrix}).

\begin{table}
\caption{Confusion Matrix}
\centering
\label{confusion_matrix}
\begin{tabular}{|c|c|c|}
\hline
\diagbox{Label}{Predict} & Positive & Negative\\ 
\hline
Positive & True Positive (TP) & False Negative (FN) \\
\hline
Negative & False Positive (FP) & True Negative (TN) \\
\hline
\end{tabular}
\end{table}

\begin{equation}
Precision = \frac{TP}{TP + FP}
\label{precision}
\end{equation}

\begin{equation}
Recall = \frac{TP}{TP + FN}
\label{recall}
\end{equation}

\begin{equation}
F1 = \frac{2 \times Precision \times Recall}{Precision + Recall}
\label{f1}
\end{equation}

\subsection{Baselines of FindVehicle}
\label{baseline_findvehicle}
 Table \ref{ner_model_results} shows the evaluation results of models on the test set of FindVehicle. It is apparent that Transformer-based models perform better than the RNN-based model. BiLSTM-CRF \citep{huang2015bidirectional} got 49.5\% F1 score, which is the lowest value among models. FLERT \citep{schweter2020flert} achieved 80.9\% F1 score, which is the highest value, 3\% higher than BERT-CRF \citep{souza2019portuguese}.

\begin{table}
\caption{Evaluation Results of Three Models on The Test Set of FindVehicle}
\centering
\label{ner_model_results}
\begin{tabular}{llll}  
\toprule   
  Model & Precision (\%) & Recall (\%) & F1 (\%)  \\
\midrule   
BiLSTM-CRF \citep{huang2015bidirectional} & 50.1 & 50.4 & 49.5 \\
BERT-CRF \citep{souza2019portuguese} & 77.7 & 78.4 & 77.9 \\
FLERT \citep{schweter2020flert} & 80.6 & 81.3 & 80.9 \\
\bottomrule  
\end{tabular}
\end{table}

Furthermore, we do the statistics on the evaluation results for all 21 classes of named entities. We take the evaluation results of FLERT \citep{schweter2020flert} as the example, as Table \ref{ner_entity_results} shows, all the evaluation metric values of fine-grained entities are much lower than those of coarse-grained entities. It denotes that the recognition of fine-grained entities is harder than coarse-grained entities for neural networks. Moreover, we also calculate the evaluation results of flat entities and overlapped entities by FLERT \citep{schweter2020flert}. As Table \ref{flat_overlapped_entity} shows, the values of three metrics of flat entity are about 20\% higher than overlapped entities'. The recognition of overlapped entities is still a challenge in FindVehicle.
\begin{table}
\caption{Evaluation Results of FLERT \citep{schweter2020flert} for All The Classes of FindVehicle}
\centering
\label{ner_entity_results}
\begin{tabular}{llll}  
\toprule   
 Entity Class & Precision (\%) & Recall (\%) & F1 (\%)  \\
\midrule   
vehicle\_color & 91.8 & 91.9 & 91.8 \\
vehicle\_brand & 91.9 & 91.8 & 91.8 \\
vehicle\_model & 91.7 & 91.7 & 91.7 \\
vehicle\_location & 91.7 & 91.8 & 91.8 \\
vehicle\_velocity & 90.4 & 89.8 & 90.1 \\
vehicle\_orientation & 91.5 & 88.7 & 90.1 \\
vehicle\_range & 92.1 & 91.9 & 92.0 \\
vehicle\_type & 91.6 & 91.8 & 91.7 \\
vehicle\_sedan & 81.9 & 83.0 & 82.4 \\
vehicle\_type-suv & 84.7 & 86.1 & 85.4 \\
vehicle\_type-motorcycle & 87.7 & 91.5 & 89.6 \\
vehicle\_type-sports\_car & 84.1 & 85.5 & 84.8 \\
vehicle\_type-hatchback & 68.8 & 66.5 & 67.6 \\
vehicle\_type-vintage\_car & 81.6 & 82.9 & 82.2 \\
vehicle\_type-coupe & 72.7 & 77.4 & 75.0 \\
vehicle\_type-truck & 74.7 & 81.0 & 77.7 \\
vehicle\_type-van & 62.9 & 74.0 & 68.1 \\
vehicle\_type-mpv & 63.1 & 66.8 & 64.9 \\
vehicle\_type-estate\_car & 62.0 & 58.7 & 60.3 \\
vehicle\_type-bus & 70.7 & 68.0 & 69.3 \\
vehicle\_type-roadster & 44.9 & 32.4 & 37.8 \\
\bottomrule  
\end{tabular}
\end{table}

\begin{table}
\caption{Evaluation Results of FLERT \citep{schweter2020flert} for Flat and Overlapped Entities of FindVehicle}
\centering
\label{flat_overlapped_entity}
\begin{tabular}{llll}  
\toprule   
  Entity Type & Precision (\%) & Recall (\%) & F1 (\%)  \\
\midrule   
Flat Entity & 91.6 & 91.5 & 91.6 \\
Overlapped Entity & 71.5 & 72.6 & 71.9 \\
\bottomrule  
\end{tabular}
\end{table}

\subsection{Comparison of Models on Different NER Datasets}
We also compare the performances of models on different NER datasets (Table \ref{ner_models_different_datasets}), including CoNLL'03 (4 classes) \cite{sang2003introduction}, WNUT'17 (6 classes) \cite{derczynski2017results}, Ontonotes (18 classes) \cite{weischedel2013ontonotes} and our FindVehicle (21 classes). We use F1 score as the evaluation metric. We can see that F1 scores (Equation \ref{f1}) of three models on FindVehicle are all lower than the scores on CoNLL'03 \cite{sang2003introduction} and Ontonotes (18 classes) \cite{weischedel2013ontonotes}, which indicates that there are some challenges in our dataset to some extent.

\begin{table}
\caption{Performances of Models on Test Sets of Different NER Datasets}
\setlength\tabcolsep{3pt}
\centering
\label{ner_models_different_datasets}
\begin{tabular}{lllll}  
\toprule   
  \diagbox{Models}{F1}{Datasets} & CoLL'03 \cite{sang2003introduction} & WNUT'17 \cite{derczynski2017results} & Ontonotes \cite{weischedel2013ontonotes} & \textbf{FindVehicle (ours)}  \\
\midrule   
BiLSTM-CRF \citep{huang2015bidirectional} & 91.7 & 42.6 & 87.1 & \textbf{49.5} \\
BERT-CRF \citep{souza2019portuguese} & 93.4 & 59.8 & 92.0 & \textbf{77.9} \\
FLERT \citep{schweter2020flert} & 94.1 & 61.1 & 92.3 & \textbf{80.9} \\
\bottomrule  
\end{tabular}
\end{table}

\section{Experiments of VehicleFinder}
\label{experiments_VehicleFinder}

There are four parts in the experiments of VehicleFinder, which are CTIM, vision detector, text detector and VehicleFinder.

\subsection{Experiments of CTIM}

\begin{figure}
\centerline{\includegraphics[width=.57\textwidth]{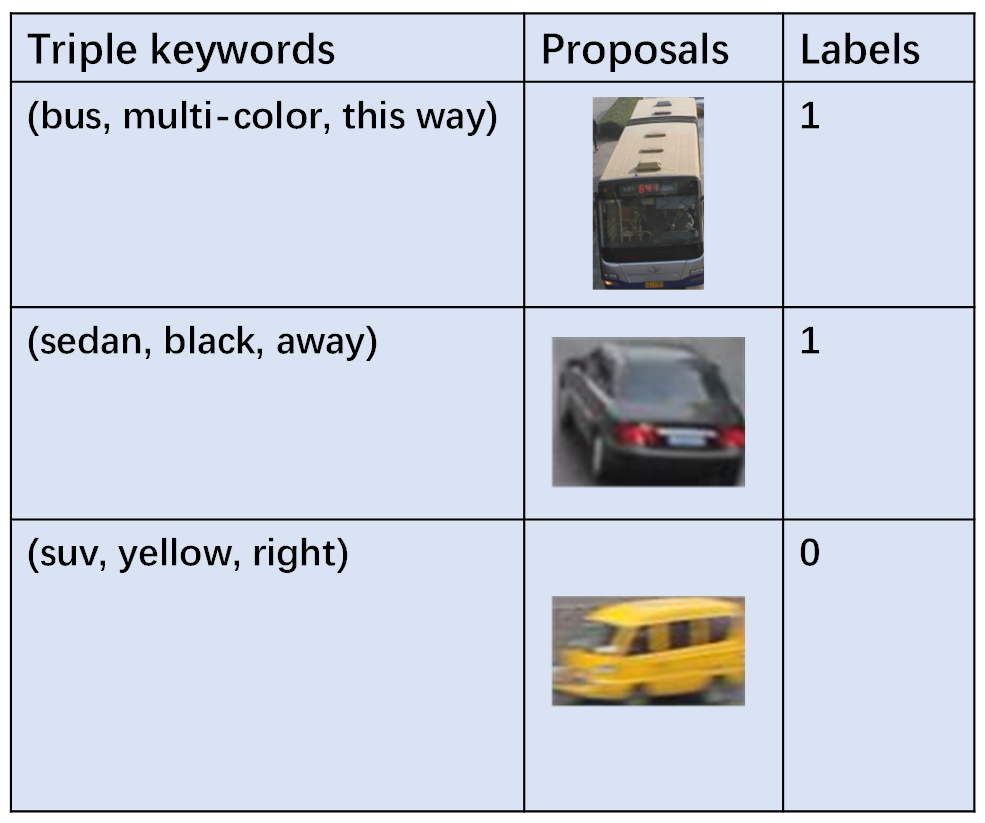}}
\caption{The data format of Vehicle-TI.}
\label{ua-detrac-ml}
\end{figure}

\subsubsection{Settings of Training and Evaluation}

We construct a text-to-image dataset called Vehicle-TI based on the training set of UA-DETRAC \citep{wen2020ua} to train and test our CTIM. As Fig. \ref{ua-detrac-ml} shows, each data sample in Vehicle-TI has a triple keyword (text modal), a proposal (image modal) and a label, which are extracted and reconstituted from UA-DETRAC \citep{wen2020ua}. A triple keyword contains the type, color and orientation of the vehicle. The label indicates whether the proposal is consistent with the description of the triple keyword, where 1 means consistent (positive sample) and 0 means inconsistent (negative sample). Positive sample is to make the feature encodings of text and image closer while the negative sample is to make the feature encodings of text and image farther. There are 598,336 samples in Vehicle-TI, 335,040 for training, 179,520 for test and 83,776 for validation.

As Table \ref{ctim_implementation} shows, we train CTIM for 50 epochs with a batch size of 64. The initial learning rate is 0.001 and CTIM is optimized by AdamW \citep{loshchilov2017decoupled}. The learning rate is scheduled by the Step scheduler.

Furthermore, we set a threshold of 0.7 as the boundary of the consistency of the vehicle proposal and the triple keyword. If the output of CTIM is above 0.7, it indicates that the vehicle proposal and the triple keyword are consistent (strong-related), if not, we think they are not related or weak-related.

\subsubsection{Evaluation Results}

Table \ref{ctim-result} presents the evaluation results of CTIM on the test set of Vehicle-TI. CTIM has only 3.84 million parameters, which gets 97.7\% accuracy (Equation \ref{accuracy}) for the identification of consistency between vehicle images and triple keywords. 

\begin{equation}
    Accuracy = \frac{TP + TN}{TP + FP + TN + FN}
\label{accuracy}
\end{equation}

Moreover, we also test the inference speed of CTIM on different devices. CTIM spends 131.42 ms identifying one sample on an 8-core ARM v8.2 of NVIDIA Jetson AGX Xavier. When tested on an i7-12700, CTIM gets 67.43 ms latency. In addition, it costs CTIM 39.47 ms on one RTX A4000. The above proves that CTIM can maintain high performance on both edge and host devices.

\begin{table}
\caption{Implementation Details of CTIM on The Training Set of Vehicle-TI}
\centering
\label{ctim_implementation}
\begin{tabular}{lllllll}  
\toprule   
  Model & Epochs & BS & ILR & Opt & Sch$^1$  \\
\midrule   
CTIM & 50 & 64 & 0.001 & AdamW & Step \\
\bottomrule  
\end{tabular}
\\
\footnotesize{$^1$ BS means batch size; ILR means initial learning rate; Opt means optimizer; Sch means scheduler.} 
\end{table}

\begin{table}
\caption{Evaluation of CTIM on The Test Set of Vehicle-TI}
\centering
\label{ctim-result}
\begin{tabular}{llllll}  
\toprule   
  Model & Param(M) & Latency(ms)          & Latency(ms)  & Latency(ms) & Accuracy(\%) \\
        &           &  8-core ARM v8.2 & i7-12700 &   RTX A4000           &       \\
\midrule   
CTIM & 3.84 & 131.42 & 67.43 & 39.47 & 97.7 \\
\bottomrule  
\end{tabular}
\end{table}

\subsubsection{Comparison of CTIM with Other Models}
As the above mention, all convolution operations in CTIM are depthwise separable convolution. In addition, we use cosine distance and linear compression function to measure and process the similarity between text and image modalities. We still call it CTIM. 

We firstly replace all depthwise separable convolution operations with the normal convolution operations, which is called CTIM-Conv-CosineD. 

Secondly, we replace the cosine distance and linear compression function in CTIM with fully connected layers, which is the operation in normal Siamese neural networks to fit the similarity by fully connected layers. We call it CTIM-DSConv-Siamese. 

Thirdly, we adopt the most well-known contrastive language and image pretraining model, CLIP \citep{radford2021learning}. We adopt ResNet-50 as the image encoder and Transformer as the text encoder. The total parameter number of CLIP is 102.58 million.

Last but not least, we fine-tune a bert-based Siamese neural network \citep{vilcek2018transformer} to make it adapt to our task. Its architecture is Transformer-based, totally different from the aforementioned neural networks. We call it Bert-Siamese.

As Table \ref{ctim-ablation} presents, all the indicators of CTIM are the best. In contrast, CTIM-Conv-CosineD-Linear has 36.6 million parameters, which is 32.83 million more than CTIM. Furthermore, its speed on different devices is slower than CTIM. 

Secondly, the parameter number of CTIM-DSConv-Siamese is the largest among all CNN-based neural networks, which is 99.72 million. Its inference speed on different devices is also the slowest and the accuracy is only 25.7\%. 

Thirdly, CLIP with ResNet-50 and Transformer has 102.58 million parameters. It spends about 2.2 seconds inferring one text-image pair sample on 8-core ARM v8.2. It gets 96.5\% accuracy on the test set.

Last but not least, although Bert-Siamese \citep{vilcek2018transformer} has a close performance with our CTIM, it has huge parameters of 189.53 million. It spends nearly 3 seconds to identify one sample on 8-core ARM v8.2, which is far too slow to deploy on edge devices.

\begin{table}
\tiny
\setlength\tabcolsep{2pt}
\caption{Ablation and Comparison Experiments on The Test Set of Vehicle-TI}
\centering
\label{ctim-ablation}
\begin{tabular}{llllll}  
\toprule   
  Model & Param(M) & Latency(ms)          & Latency(ms)  & Latency(ms) & Accuracy(\%) \\
        &           &  8-core ARM v8.2 & i7-12700 &   RTX A4000           &       \\
\midrule   
CTIM & 3.84 & 131.42 & 67.43 & 39.47 & 97.7 \\
CTIM-Conv-CosineD-Linear & 36.67 & 185.64 & 79.72 & 55.54 & 97.2 \\
CTIM-DSConv-Siamese       & 99.72 & 211.79 & 134.45 & 96.18 & 25.7 \\
CLIP(ResNet-50+Transformer)  & 102.58 & 2236.12 & 1395.77 & 527.86 & 96.5 \\
Bert-Siamese  & 189.53 & 2819.87 & 1657.53 & 886.62 & 97.3 \\
\bottomrule  
\end{tabular}
\end{table}

\subsection{Experiments of Vision Detector}
Vision detector is to extract proposals of vehicles from the image. We adopt NanoDet-m \citep{nanodet} as the vision detector, which is a lightweight detector with only 0.95 million parameters. It is trained on the training set of UA-DETRAC \citep{wen2020ua}. The implementation details are shown in Table \ref{nanodet-m}. 

Moreover, we want the vision detector to miss as few targets as possible, so we use recall as the evaluation metric instead of precision. As Table \ref{nanodet-m-result} presents, NanoDet-m \citep{nanodet} gets 86.7\% recall rate on the test set. 

\begin{table}
\caption{Implementation Details of NanoDet-m on The Training Set of UA-DETRAC \citep{wen2020ua}}
\centering
\label{nanodet-m}
\begin{tabular}{llllll}  
\toprule   
  Model & Epochs & BS & ILR & Opt & Sch$^1$  \\
\midrule   
NanoDet-m \citep{nanodet} & 200 & 16 & 0.001 & AdamW & Cosine \\
\bottomrule  
\end{tabular}
\\
\footnotesize{$^1$ BS means batch size; ILR means initial learning rate; Opt means optimizer; Sch means scheduler.} 
\end{table}

\begin{table}
\tiny
\caption{Evaluation of NanoDet-m on UA-DETRAC \citep{wen2020ua}}
\centering
\label{nanodet-m-result}
\begin{tabular}{lllllll}  
\toprule   
  Model & Param(M) & Latency(ms)          & Latency(ms)  & Latency(ms) & Recall(\%)\\
        &           &  8-core ARM v8.2 & i7-12700 &  RTX A4000 &                   & \\
\midrule   
NanoDet-m \citep{nanodet} & 0.95 & 7.99 & 3.72 & 1.14 & 86.7\\
\bottomrule  
\end{tabular}
\end{table}

\subsection{Experiments of Text Detector}
Text detector is to extract keywords (named entities) from the user command. BiLSTM-CRF has relatively few parameters and fast inference among all NER models mentioned in Section \ref{baseline_findvehicle}, so we train a BiLSTM-CRF on our FindVehicle as the text detector, which is to extract named entities with types of \textit{vehicle\_type}, \textit{vehicle\_color} and \textit{vehicle\_orientation}. The implementation details are shown in Table \ref{ner_model_hyper}.

As Table \ref{bilstm-crf-result1} shows, BiLSTM-CRF has 4.02 million parameters. It spends 148.57 ms extracting all named entities from a sample in FindVehicle on the 8-core ARM v8.2. In addition, it spends 87.19 ms and 51.73 ms when tested on i7-12700 and RTX A4000.

\begin{table}
\caption{Inference Speed Evaluation of BiLSTM-CRF \citep{huang2015bidirectional}}
\centering
\label{bilstm-crf-result1}
\begin{tabular}{lllll}  
\toprule   
  Model & Param(M) & Latency(ms)          & Latency(ms) & Latency(ms) \\
        &          & 8-core ARM v8.2      &  i7-12700 &  RTX A4000\\
\midrule   
BiLSTM-CRF \citep{huang2015bidirectional} & 4.02 & 148.57 & 87.19 & 51.73\\
\bottomrule  
\end{tabular}
\end{table}

\begin{table}
\caption{F1 Scores of Different Kinds of Named Entities by BiLSTM-CRF \citep{huang2015bidirectional} on FindVehicle}
\centering
\label{bilstm-crf-result2}
\begin{tabular}{llll}  
\toprule   
  Model & F1 (\textit{vehicle\_type})  & F1 (\textit{vehicle\_color}) & F1 (\textit{vehicle\_orientation})\\
\midrule   
BiLSTM-CRF \citep{nanodet} & 90.56 & 89.79 & 90.17\\
\bottomrule  
\end{tabular}
\end{table}

\subsection{Evaluation of VehicleFinder}

\subsubsection{Settings of Evaluation}
\begin{figure}
\centerline{\includegraphics[width=.90\textwidth]{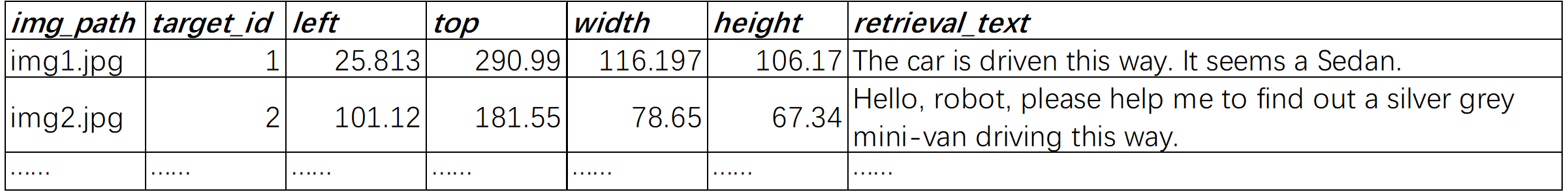}}
\caption{The format of the homemade test set for VehicleFinder.}
\label{home_made}
\end{figure}

We randomly sample 2000 images from the test set of UA-DETRAC \citep{wen2020ua} as our homemade test set for VehicleFinder. For each image, we write a piece of retrieval text, which corresponds to one or more vehicles in the image. The format of the test set is presented in Fig. \ref{home_made}. Each item includes columns of the image path \textit{img\_path}, target id \textit{target\_id}, the upper-left abscissa of bounding box \textit{left}, the upper-left ordinate of bounding box \textit{top}, the width of bounding box \textit{width}, the height of bounding box \textit{height} and the retrieval content \textit{retrieval\_text}. There are 3917 target vehicles based on retrieval text in these 2000 images. We adopt precision, recall and F1 score to evaluate our VehicleFinder, which are presented in Equation \ref{precision_vehicle_finder1}, \ref{precision_vehicle_finder2} and \ref{precision_vehicle_finder3}. We also test our VehicleFinder on three different devices.

\begin{align}
\label{precision_vehicle_finder1}
& Precision_{V} = \frac{num(\textit{detected vehicles} \textbf{ \& } \textit{detected vehicles in the testset})}{\textit{num(detected vehicles)}}
\end{align}
\begin{align}
\label{precision_vehicle_finder2}
& Recall_{V} = \frac{num(\textit{detected vehicles} \textbf{ \& } \textit{detected vehicles in the testset})}{\textit{num(all vehicles in the testset)}}
\end{align}
\begin{align}
\label{precision_vehicle_finder3}
& F1 = \frac{2 \times Precision_{V} \times Recall_{V}}{Precision_{V} + Recall_{V}}
\end{align}

\subsubsection{Evaluation Results}
\begin{figure}
\centerline{\includegraphics[width=.96\textwidth]{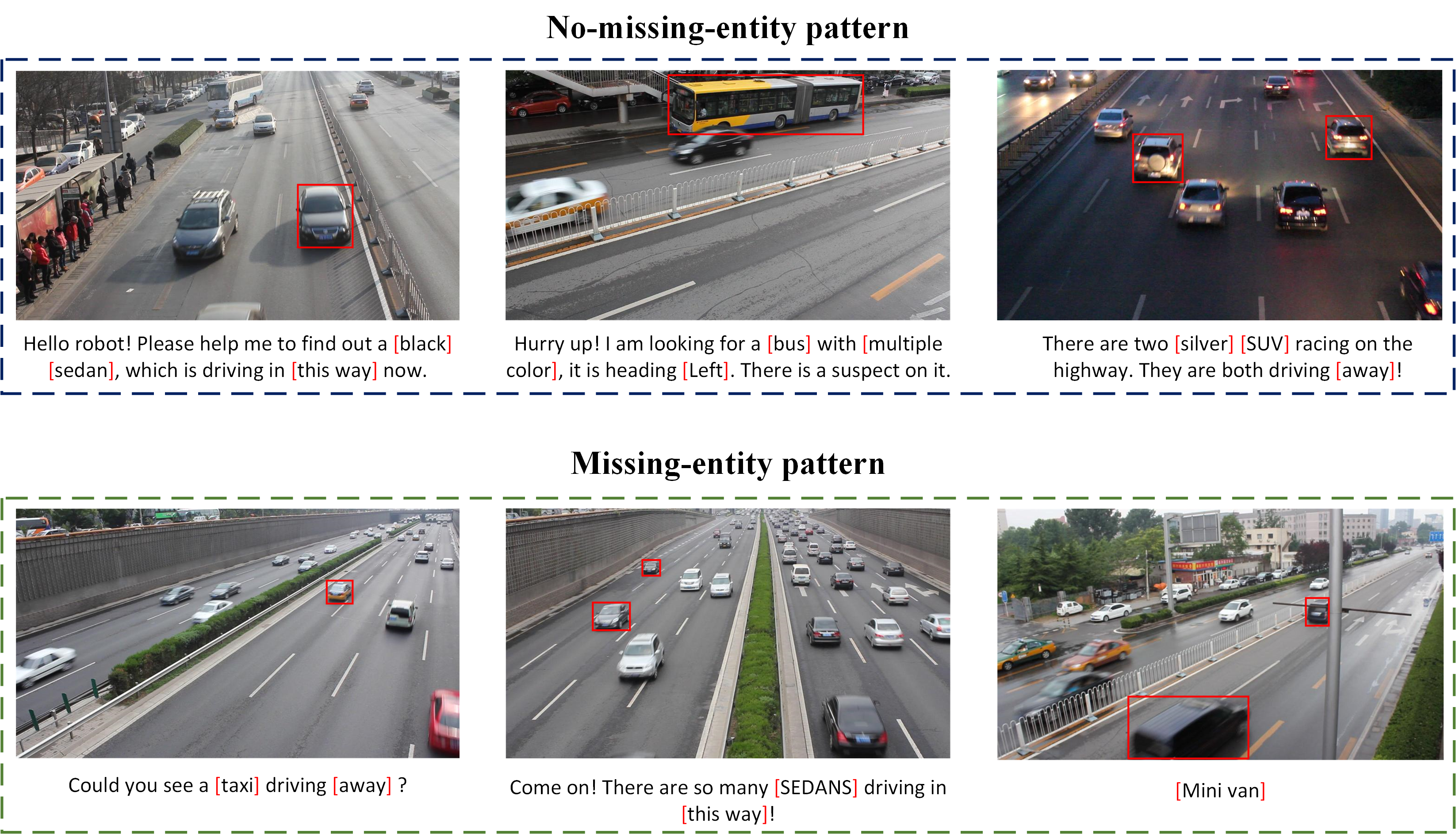}}
\caption{Samples of inference results by VehicleFinder on UA-DETRAC \citep{wen2020ua}.}
\label{inference_result_vehiclefinder}
\end{figure}

Table \ref{vehiclefinder_f1} shows that our VehicleFinder(CTIM) has 8.81 million parameters, containing the vision detector, the text detector and CTIM. After setting two thresholds $th_{nm}$ and $th_{m}$ as 0.70 and 0.30 respectively, our VehicleFinder(CTIM) achieves 87.7\% precision, 89.4\% recall and 88.5\% F1 score. Fig. \ref{inference_result_vehiclefinder} presents the test results of VehicleFinder(CTIM). We can observe that the targeted vehicles could be preciously retrieved based on the description. 

Furthermore, we also collect the results of the control group. VehicleFinder(CTIM-Conv-CosineD-Linear) achieves 87.4\% precision, 87.9\% recall and 87.6\% F1 score with 41.64 million parameters. VehicleFinder(CTIM-DSConv-Siamese) has 104.69 million parameters and its F1 score is only 12.5\%, which is the lowest among all. VehicleFinder(CLIP) has 107.55 million parameters total and gets 87.5\% F1 score. VehicleFinder(Bert-Siamese) has the almost same F1 score (88.4\%) as our VehicleFinder(CTIM), but its parameters are too huge.

We calculate the latency of our VehicleFinder(CTIM) from the moment that the command is loaded into VehicleFinder(CTIM) to the moment that the VehicleFinder(CTIM) completes the identification of one vehicle. As Equation \ref{vehiclefinder_latency} presents, $T_{ner}$ means the time of named entity recognition and $T_{cti}$ means the identification time of the consistency of named entities and one vehicle proposal. It includes the inference time of the text detector and CTIM. We ignore the time for the system to schedule different models. Table \ref{vehiclefinder_speed} shows the inference speed evaluation of our VehicleFinder(CTIM). The longest latency is 279.35 ms on one 8-core ARM v8.2 while the shortest latency is 93.72 ms on one RTX A4000. It implies that our VehicleFinder(CTIM) could be deployed on both edge devices and host devices, but host devices are the better choice.

In contrast, firstly, VehicleFinder(CTIM-Conv-CosineD-Linear) is 86.42 ms slower than VehicleFinder(CTIM) on one ARM v8.2 CPU and 40.85 ms slower on one RTX A4000 GPU. Secondly, VehicleFinder(CTIM-DSConv-Siamese) is 119.05 ms slower than VehicleFinder(CTIM) on one ARM v8.2 CPU and 73.73 ms slower on one RTX A4000 GPU. Thirdly, VehicleFinder(CLIP) spends above 2 seconds inferring one text-image pair on ARM v8.2 CPU. Fourthly, VehicleFinder(Bert-Siamese) has the slowest inference among all, whose latency on one ARM v8.2 CPU exceeds 3 seconds (3091.33 ms) and almost 1 second on one RTX A4000 GPU. The inference speed of VehicleFinder(Bert-Siamese) is far too slow to deploy on no matter edge devices or host devices to use in actual traffic scenes.

\begin{figure}
\centerline{\includegraphics[width=.96\textwidth]{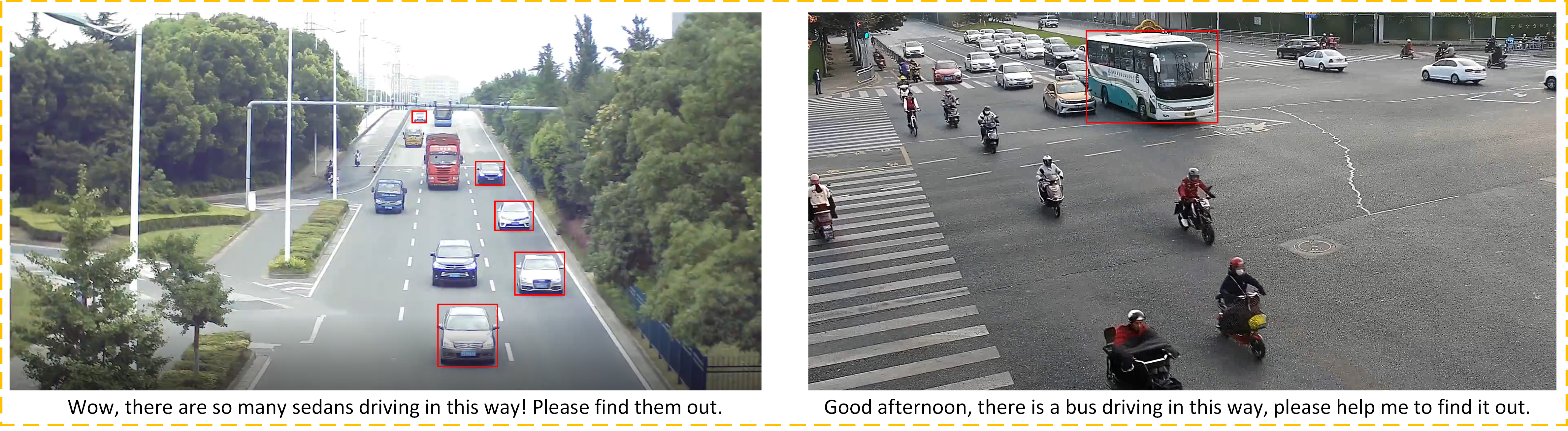}}
\caption{Inference results of VehicleFinder on our collected images.}
\label{inference_result_vehiclefinder2}
\end{figure}

\begin{figure}
\centerline{\includegraphics[width=.96\textwidth]{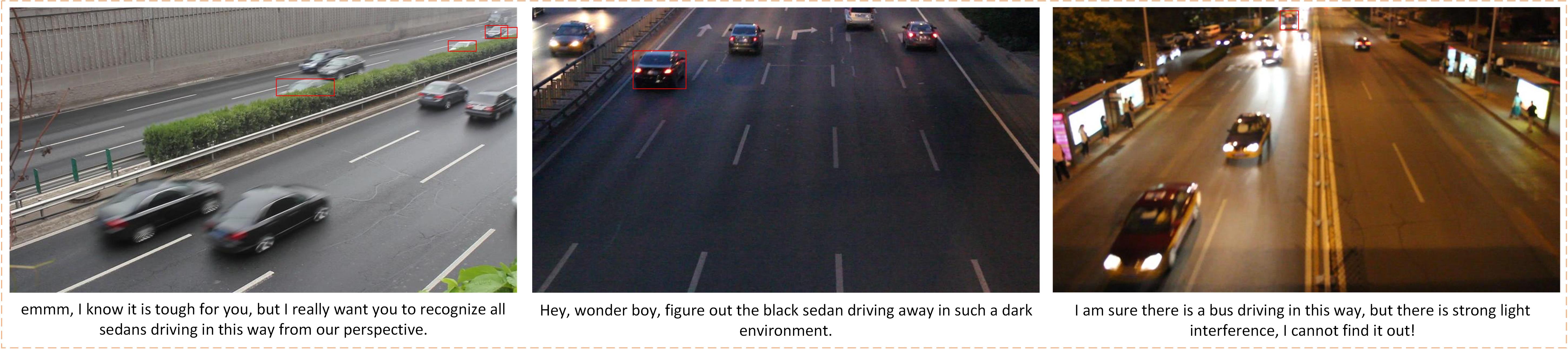}}
\caption{Inference results of corner cases: occluded targets, dark environment and strong light interference.}
\label{adverse_samples}
\end{figure}

Last but not least, we also test the VehicleFinder(CTIM) on our collected images in some traffic scenes. Based on the images, we recruited two volunteers to describe the vehicles that they want to find out in the images. As Fig. \ref{inference_result_vehiclefinder2} presents, VehicleFinder(CTIM) can still accurately find out the targeted vehicles based on the volunteers' descriptions. In addition, inference of corner cases is included as Fig. \ref{adverse_samples} shows, which means VehicleFinder(CTIM) can keep robust to some extent when confronted with some adverse phenomenons.

\begin{equation}
    Latency = T_{ner} + T_{cti}
\label{vehiclefinder_latency}
\end{equation}

\begin{table}
\tiny
\caption{Evaluation of VehicleFinder on Our Homemade Test Set}
\centering
\label{vehiclefinder_f1}
\begin{tabular}{lllll}  
\toprule   
  Model & Param(M) & Precision(\%) & Recall(\%) & F1(\%)\\
\midrule   
VehicleFinder(CTIM) & 8.81 & 87.7 & 89.4 & 88.5\\
VehicleFinder(CTIM-Conv-CosineD-Linear) & 41.64 & 87.4 & 87.9 & 87.6 \\
VehicleFinder(CTIM-DSConv-Siamese) & 104.69 & 11.7 & 13.4 & 12.5 \\
VehicleFinder(CLIP) & 107.55 & 86.6 & 88.4 & 87.5 \\
VehicleFinder(Bert-Siamese) & 194.50 & 87.5 & 89.4 & 88.4 \\
\bottomrule  
\end{tabular}
\end{table}

\begin{table}
\tiny
\caption{Inference Speed Evaluation of VehicleFinder}
\centering
\label{vehiclefinder_speed}
\begin{tabular}{lllll}  
\toprule   
  Model & Param(M) & Latency(ms)          & Latency(ms) & Latency(ms) \\
        &          & 8-core ARM v8.2      &  i7-12700 &  RTX A4000\\
\midrule   
VehicleFinder(CTIM) & 8.81 & 279.35 & 169.34 & 97.72\\
VehicleFinder(CTIM-Conv-CosineD-Linear) & 41.64 & 365.77 & 212.59 & 138.57 \\
VehicleFinder(CTIM-DSConv-Siamese) & 104.69 & 398.35 & 243.93 & 171.45 \\
VehicleFinder(CLIP) & 107.55 & 2398.77 & 1487.53 & 588.56\\
VehicleFinder(Bert-Siamese) & 194.50 & 3094.33 & 1809.51 & 973.83 \\
\bottomrule  
\end{tabular}
\end{table}

\section{Conclusion and Future Work}
\label{conclusion_future}

We propose the first NER dataset FindVehicle in traffic domain, which contains different sentences that describe the vehicles in different traffic scenes. Named entities include several attributes of vehicles that could be detected by perception sensors. FindVehicle is a NER dataset that contains both flat and overlapped entities. All the named entities in it are annotated by both machine annotation algorithms and human annotators. Annotation includes both coarse-grained and fine-grained entity annotation. FindVehicle could be used to assist text-image cross-modal tasks in traffic scenes and act as the pretrained corpus of the territory of traffic. Furthermore, We propose an efficient text-image cross-modal vehicle retrieval system called VehicleFinder. VehicleFinder achieves 87.7\% precision when identifying target vehicles by text commands, which spends 279.35 ms on one 8-core ARM v8.2 CPU and 93.72 ms on one RTX A4000 GPU. Our VehicleFinder could help traffic supervisors find out the target vehicle from a large number of images or videos based on natural language. Last but not least, we construct a text-to-image vehicle-matching dataset called Vehicle-TI.

In the future, firstly, we will continue to maintain our FindVehicle. Secondly, we will extend FindVehicle by adding the corpus of some special traffic scenes, and connecting samples of FindVehicle to images of real traffic scenes, which would be a new dataset (benchmark). Thirdly, we will explore text-video cross-modal vehicle retrieval.

\section{Discussion}
\label{discussion}

The discussion is divided into two parts, the challenges of FindVehicle and the limitation of our cross-modal vehicle retrieval system VehicleFinder.

\begin{figure}
\centerline{\includegraphics[width=.57\textwidth]{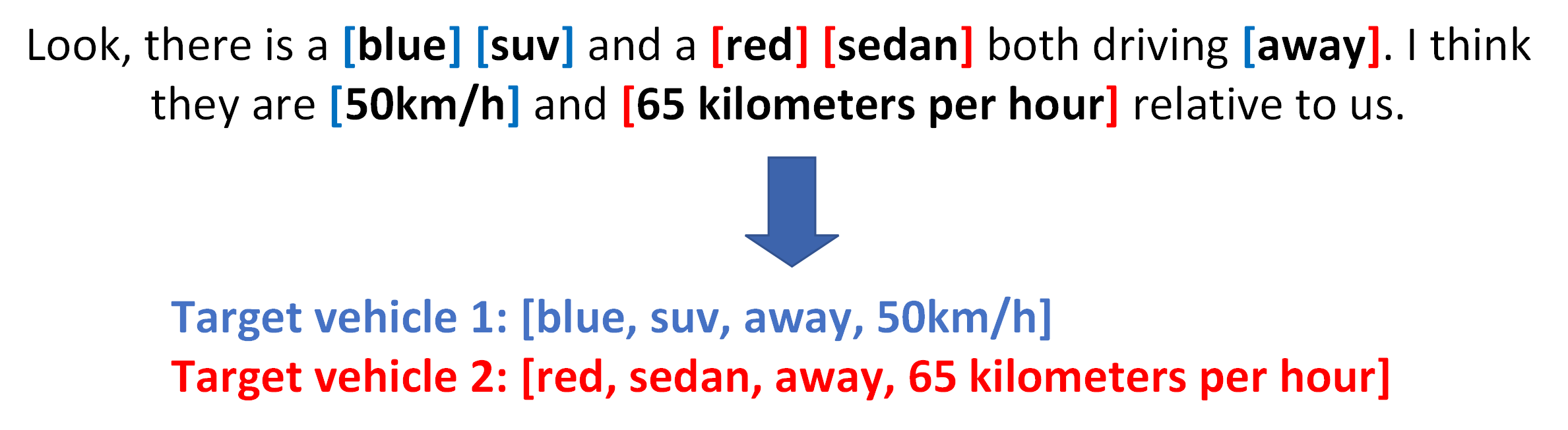}}
\caption{The challenge of multiple entity clustering.}
\label{relation_extraction}
\end{figure}

In FindVehicle, long-tail data distribution, the recognition of vehicle brands out of the distribution, and the recognition of fine-grained and overlapped entities are three challenges worth exploring. Moreover, as Fig. \ref{relation_extraction} shows, identifying whether the extracted named entities refer to the same vehicle is a considerable challenge, equivalent to clustering named entities according to context.

The first limitation of our VehicleFinder is that a description can only contain the attributes of one vehicle. Our VehicleFinder is not adaptive to context with multiple vehicles because we only adopt NER in keyword extraction instead of combining NER with relation extraction, which is also a challenge in the future. The second limitation is that the granularity of the keywords used to describe the attributes of the vehicles is not fine enough, which is attributed to the limitation of the human cost of the annotation effort. We will continue to pay attention to and research this field in the future.

\backmatter

\bmhead{Acknowledgments}

The authors acknowledge XJTLU-JITRI Academy of Industrial Technology for giving valuable support to the joint project. This work is also partially supported by the Xi’an Jiaotong-Liverpool University (XJTLU) AI University Research Centre, Jiangsu (Provincial) Data Science and Cognitive Computational Engineering Research Centre at XJTLU (funding: XJTLU-REF-21-01-002). The authors sincerely acknowledge Sihao Dai, Zhou Yuan, Wenjie Zhou for their help in the project.

\section*{Declarations}

\begin{itemize}
\item \textbf{Availability of data and materials}: All the data are available to access in the first author's GitHub repository.
\item \textbf{Code availability}: All the code are available to access in the first author's GitHub repository.
\item \textbf{Conflict of Interest}: The authors declare that they have no conflict of interest.
\end{itemize}

\bibliography{sn-bibliography}


\end{document}